\documentclass[10pt,journal,compsoc]{IEEEtran}
\usepackage{graphicx}
\usepackage{amsfonts}
\usepackage{diagbox}
\usepackage{multirow}
\usepackage{graphics}
\usepackage{epstopdf}
\usepackage{tabularx}
\usepackage{booktabs}
\usepackage{epsfig}
\usepackage{pifont}
\usepackage{color,subfigure}
\usepackage{mathrsfs}
\usepackage{bm}
\usepackage{algorithm, algorithmic}
\usepackage{mathdots}
\usepackage[american]{babel}
\usepackage{array}
\usepackage{setspace}
\usepackage{threeparttable}
\usepackage{amsthm,amsmath,amssymb}
\usepackage{url}
\usepackage[colorlinks,linkcolor=red]{hyperref}

\ifCLASSOPTIONcompsoc
  \usepackage[nocompress]{cite}
\else
  \usepackage{cite}
\fi

\ifCLASSINFOpdf
\else
\fi

\begin{document}

\title{Efficient Robustness Assessment via \\ Adversarial Spatial-Temporal Focus on Videos}

\author{Xingxing Wei, \emph{Member, IEEE},
        Songping Wang,
        and~Huanqian Yan
\IEEEcompsocitemizethanks{
\IEEEcompsocthanksitem Xingxing Wei, Songping Wang  and Huanqian Yan are with the Institute of Artificial Intelligence, Beihang University, No.37, Xueyuan Road, Haidian District, 100191, Beijing, P.R. China. 
\IEEEcompsocthanksitem Xingxing Wei is the corresponding author (xxwei@buaa.edu.cn)}
}

\markboth{IEEE TRANSACTIONS ON PATTERN ANALYSIS AND MACHINE INTELLIGENCE}%
{Shell \MakeLowercase{\textit{et al.}}: Bare Demo of IEEEtran.cls for Computer Society Journals}

\IEEEtitleabstractindextext{%
\begin{abstract}
Adversarial robustness assessment for video recognition models has raised  concerns owing to their wide applications on safety-critical tasks. Compared with images, videos have much high dimension, which brings huge computational costs when generating adversarial videos. This is especially  {serious} for the query-based black-box attacks  {where} gradient estimation for the threat models is usually utilized, and high dimensions will lead to a large number of queries.  To mitigate this issue,  we propose to \textbf{simultaneously} eliminate the temporal and spatial redundancy within the video to achieve an effective and efficient gradient estimation on the reduced  searching space, and thus query number could decrease. To implement this idea, we design the novel \textbf{A}dversarial \textbf{s}patial-\textbf{t}emporal \textbf{Focus}  (\textbf{AstFocus}) attack on videos, which performs attacks on the simultaneously focused  key frames and key regions  from the inter-frames and intra-frames in the video.  AstFocus attack is based on the cooperative Multi-Agent Reinforcement Learning (MARL) framework. One agent is responsible for selecting key frames, and another agent is responsible for selecting key regions. These two agents are jointly trained by the common rewards received from the black-box threat models to perform a cooperative prediction. By continuously querying, the reduced searching space composed of key frames and key regions is becoming precise, and the whole query number becomes less than that on the original video.  Extensive experiments on  {four} mainstream video recognition models and  {three} widely used action recognition datasets demonstrate that the proposed AstFocus attack outperforms the SOTA methods, which is prevenient in fooling rate, query number, time, and perturbation magnitude at the same.
\end{abstract}

}

\maketitle

\IEEEdisplaynontitleabstractindextext

\IEEEpeerreviewmaketitle

\IEEEraisesectionheading{\section{Introduction}\label{sec:introduction}}
\IEEEPARstart{D}{eep} Neural Networks (DNNs) have made remarkable achievements in various tasks such as object detection \cite{wang2021salient}, action recognition \cite{zhu2020comprehensive}, scene understanding \cite{yang2018scene}, and so on. Recent studies illustrate the DNNs' vulnerability to the so-called adversarial examples \cite{goodfellow2014explaining, ilyas2019adversarial, moosavi2017universal}. Afterwards, a series of methods are proposed to evaluate the adversarial robustness of DNNs. Among these works, the attack-based robustness evaluation methods \cite{wang2021understanding, tang2021robustart, geisler2021robustness} are more popular  and practical because of their good implementability. They mainly seek for the minimum adversarial perturbations of successful attacks to measure the robustness \cite{chapman2021fimap}. On one hand, accurate assessment for adversarial robustness can help to deploy DNNs into safety-critical systems. On the other hand, it provides a quantitative metric to design more robust DNNs. Therefore, adversarial robustness assessment is important in both theoretical and practical values.

Video recognition \cite{hara2018can, wang2016temporal, lin2019tsm} is a major branch in computer vision. Leveraging the temporal and spatial relationship within the video data can effectively locate and classify the objects or behaviors in videos, and thus help to perform video analysis. Owing to the DNNs' advantage, current video recognition models are usually designed based on DNNs.  The DNNs' vulnerability is inevitably inherited by video recognition models. Owing to the wide applications in some safety-critical tasks like security surveillance, evaluating their  adversarial robustness becomes necessary. Currently, more and more users begin to employ the video recognition APIs released by commercial cloud platforms  because of their easy accessibility. In such cases, the APIs' details are not public, we can only assess their adversarial robustness according to the outputs obtained by  querying the systems. So these methods are called as query-based black-box attacks, which mainly rely on the estimated gradients for the APIs \cite{dong2021query,  wei2022adversarial}.

 Compared with images, videos have much high dimensions owing to the additional temporal information, which brings huge computational costs when generating adversarial videos. This is especially  {serious} for the query-based black-box attacks because the high-dimension video data needs a large number of queries to obtain an accurate gradient estimation. Thus, seeking for the minimum adversarial perturbations on videos is more challenging than that on images, a reasonable attack algorithm should reduce the video dimensions firstly, so as to improve the attacks' efficiency and reduce the perturbations' magnitude.  {To meet this goal, temporally sparse video attacks \cite{wei2019sparse,wei2022sparse,hwang2021just} are proposed to eliminate the redundancy in the temporal domain, and 
 spatial video attacks \cite{jiang2019black} try to eliminate the redundancy in the spatial domain.  More importantly,  the spatial and temporal redundancy should be jointly considered, i.e,  modeling the key regions within key frames, and then evaluating the  robustness on these areas. The current related methods \cite{wei2019heuristic,wang2021reinforcement} both regard the selecting key frames and selecting key regions as two separate steps, and don't simultaneously consider their interaction, thus leading to the sub-optimal attacking efficiency and performance. }

\begin{figure*}[t]
\begin{center}
\includegraphics[width=0.9\linewidth]{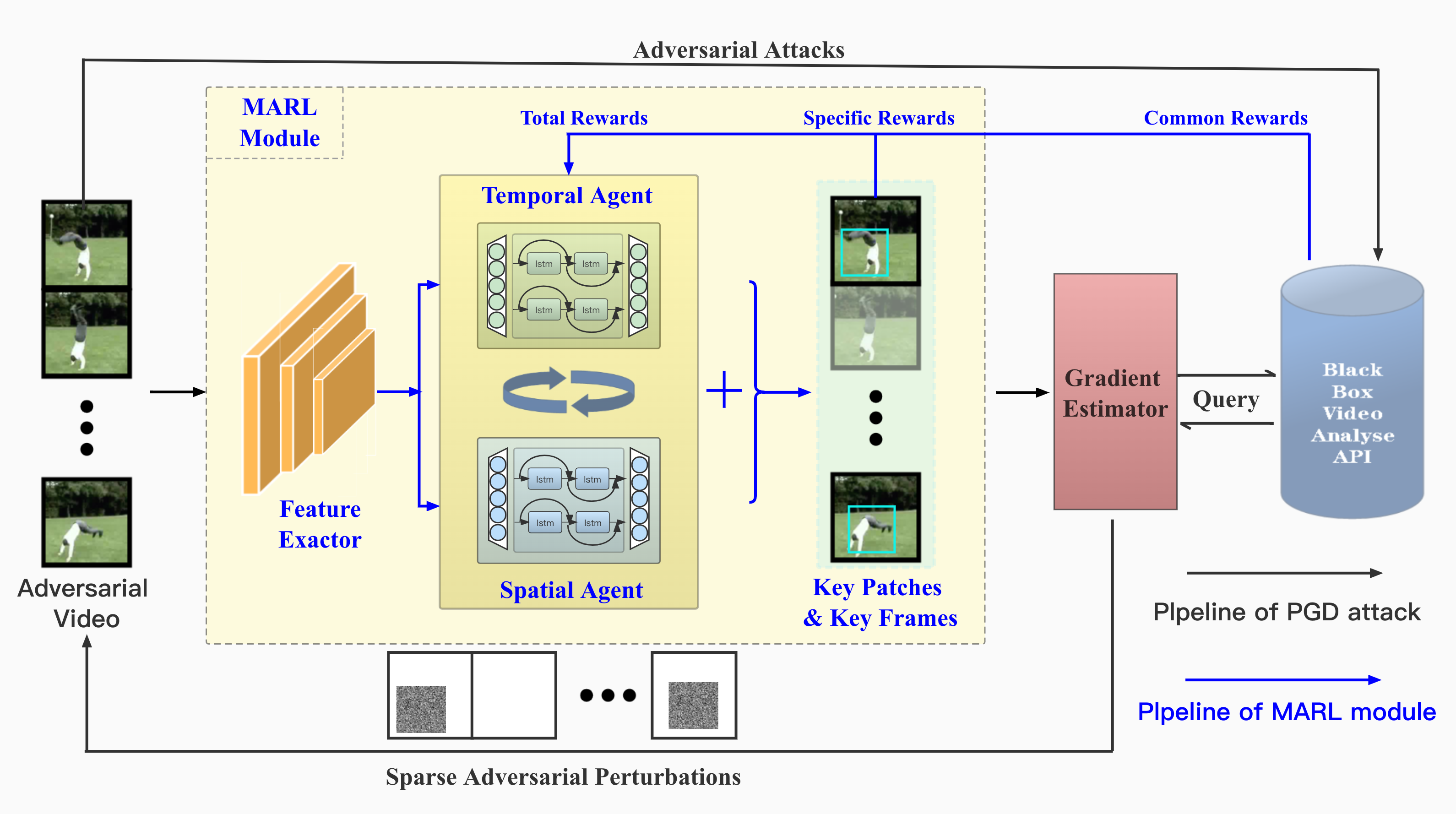}
\end{center}
\vspace{-0.5cm}
\caption{Overview of the proposed AstFocus attack. It integrates a cooperative Multi-Agent Reinforcement Learning (MARL) module into the PGD attack with NES gradient estimation  \cite{ilyas2018black}, and thus selects key frames and key patches within the video to reduce dimensions. In this way, an effective and efficient gradient estimation on the reduced  space is achieved, and the evaluation's efficiency and accuracy are improved at the same.}
\label{fig:framework}
\end{figure*}

However, simultaneously optimizing the key frames and key regions is difficult. Because they belong to different domains, and are closely coupled, i.e, changing the key frames also affects the selection of key regions. This is more challenging in the query-based black-box attacks, where  only the feedback from the threat model can be used to perform the optimization. Considering the above points, this paper mainly addresses the following problem: 
\emph{How to simultaneously learn the precise key frames and key regions to efficiently and accurately assess the adversarial robustness of video recognition in the query-based black-box setting?}

To answer this question, in this paper,  we design the novel \textbf{A}dversarial \textbf{s}patial-\textbf{t}emporal \textbf{Focus} (\textbf{AstFocus}) attack on videos, which performs attacks on the simultaneously focused  key frames and key regions  from the inter-frames and intra-frames in the video.  The key frames and key regions are dynamically adjusted by the interaction with the threat model. Technically, this process is achieved  based on the cooperative Multi-Agent Reinforcement Learning (MARL) \cite{lowe2017multi}. One agent is responsible for selecting key frames (temporal agent), and another agent is responsible for selecting key regions (spatial agent).  {These two agents use one backbone network, and are jointly trained by the common rewards received from the black-box threat models to perform a cooperative prediction.} By continuously querying, the focused space composed of key frames and key regions is becoming precise, and the whole query number becomes less than that on the original video. 

More specifically, AstFocus attack is constructed based on the PGD+NES baseline, which extends PGD \cite{madry2017towards} to the black-box attack with Natural Evolution Strategy (NES) \cite{wierstra2008natural} gradient estimator.   {We attach two agents before the gradient estimator module to reduce the video dimension.} In each PGD iteration, NES gradient estimator is first performed on the selected key frames and key regions predicted by agents. Then the local adversarial perturbations are generated to attack the threat model. Finally, these two agents are updated according to the computed rewards to predict better key frames and key regions in the next iteration. This process is continuously repeated until the successful attack is achieved. 
  These two agents have similarities and also differences.  {For policy networks, we apply the same one backbone network to extract the feature maps from the input video frames for both of them}, but design the distinct LSTM-based \cite{greff2016lstm} structures according to their own characteristic to predict the optimal actions. For actions, the temporal agent's actions are defined as the sets composed of different key frames. For the spatial agent, actions are defined as the sets composed of different patch regions located in each frame. For rewards, three rewards are carefully designed to train the agents.  {The first one is the common reward from the feedback of black-box threat models, which is used to simultaneously guide two agents.} The following two rewards are specially designed for temporal and spatial agent, respectively. And they mainly measure the actions from the view of appearance. The whole flowchart of AstFocus attack is shown in Figure \ref{fig:framework}, and the code is released in \url{https://github.com/DeepSota/AstFocus}.

This paper is an extended work based on our conference version \cite{yan2021efficient} and has the following major improvements. Firstly, we consider the spatial redundancy besides the temporal redundancy in the previous version, and further propose the novel AstFocus attack to simultaneously learn key frames and key regions and generate perturbations. This is a major change in the idea, which comprehensively makes use of the videos' spatial-temporal character to perform attacks.  Secondly, we design a cooperative multi-agent RL based method to implement the new idea, while the previous version uses single-agent RL. Thus the rewards, actions, and policies are carefully re-designed. Thirdly, more experiments are given and discussed involving the parameter tuning, ablation study, and comparisons with SOTA methods. We also re-write the abstract, introduction,  methodology, and experiment sections to better introduce our motivation and methods. We believe these modifications can significantly improve the quality of our work.

In summary, this paper has the following contributions:
\begin{itemize}
\item We propose AstFocus attack, a novel query-based black-box attack method to assess the adversarial robustness for video recognition models, the adversarial perturbations are only added on the key spatial-temporal focused spaces, which can help reduce attack query numbers and perturbations significantly.
\item A cooperative multi-agent reinforcement learning module is adopted for identifying the key frames and key regions at the same. For that, we carefully design the actions, policy networks, and rewards for both the agents according to the specific task. The agents are updated in each iteration rather than after each round of successful attack, so is efficient to converge.
\item Compared with the state-of-the-art video attack algorithms, the proposed AstFocus attack can achieve less query number and smaller adversarial perturbations. Specifically, it reduces at least 10\% query number, and improves at least 5\% fooling rate with the smallest perturbations, which verifies the efficiency and effectiveness of AstFocus attacks.
\end{itemize}

The rest of this paper is organized as follows: we briefly review the related works  in Section 2. The proposed AstFocus attack algorithm is described in Section 3. Experimental results and analysis are presented in Section 4. Finally, we conclude the whole paper in Section 5.

\section{Related Works}

\subsection{Adversarial Attacks on Videos}
Adversarial example \cite{goodfellow2014explaining, madry2017towards, yuan2019adversarial} is a maliciously crafted input designed for making the classifier produce wrong output. To make human imperceptible of its existence, the generation of adversarial examples is often limited by some deliberate conditions, such as noise size and query numbers. Adversarial video attack and adversarial image attack are similar, the difference is that the attack space of the video is much larger than that of images. It is not easy to directly extend some image attack algorithms to attack such high-dimension video data. High dimensions usually bring huge  search space, leading to high costs to achieve successful attacks. Especially in the black-box setting, a huge search space will bring  {a large number of queries}. 

Some video attack techniques have been proposed to find adversarial videos. Wei \textit{et al}. \cite{wei2019sparse} generate sparse 3D adversarial perturbations to add on the whole video. To reduce the attacking space, an $l_{2,1}$-norm regularization based optimization is designed for making the adversarial perturbations more concentrated in some key frames of the input video. This method shows the sparse ability of adversarial video noises. Similarly, \cite{hwang2021just} propose ``one frame attack'', they only add adversarial noise on one video frame. The perturbation can easily defeat deep learning-based action recognition systems. The vulnerable frame is perturbed with a gradient-based adversarial attack method. In addition, \cite{li2018adversarial} finds that the temporal structure is key to generating adversarial videos. They have used generative adversarial network to generate adversarial examples that can cause large misclassification rate for the video recognition models.
\begin{table}[t]
\caption{Comparisons with query-based black-box video attack methods. ``Temporal" denotes  reducing temporal redundancy in the video, ``Spatial" denotes reducing spatial redundancy, ``Jointly" denotes jointly learning for reducing the spatial and temporal redundancy in the video. }
\centering
\scalebox{0.95}{
\begin{tabular}{c|c|c|c}
\hline
& Temporal  &  Spatial  & Jointly \\
\hline
VBAD attack \cite{jiang2019black} &  \ding{56} &   \ding{52}&  \ding{56} \\
\hline
Sparse attack \cite{wei2022sparse} & \ding{52} &  \ding{56} &  \ding{56} \\
\hline
Motion-sampler attack \cite{zhang2020motion} & \ding{52} &  \ding{56} &  \ding{56} \\
\hline
GEO-TRAP attack \cite{li2021adversarial}  & \ding{52} &  \ding{56} &  \ding{56} \\
\hline
Heuristic attack \cite{wei2019heuristic} &  \ding{52} &   \ding{52} &  \ding{56} \\
\hline
 {RLSB attack} \cite{wang2021reinforcement}  & {\ding{52}} & {\ding{52}} &   {\ding{56}} \\
\hline
AstFocus attack (ours) & \ding{52} &  \ding{52} & \ding{52} \\
\hline
\end{tabular}}
\label{tab:trans}
\vspace{-0.3cm}
\end{table}

Not only white-box video attacks, but also black-box video attacks are explored.  {One class of such methods is based on transferablity across different models. For example, Wei \textit{et al}.\cite{cross2022cvpr} perform black-box video attacks based on  adversarial perturbations generated on image models. }

Another black-box video attacks belong to query-based methods. They generate perturbations via querying the target video recognition system. Among them, Jiang \textit{et al}.\cite{jiang2019black} extend PGD algorithm to video attack with gradient estimators computed using super-pixels.
To reduce attacking costs, some efficient black-box video attack algorithms are proposed. \cite{zhang2020motion} argues the initialized random noises in \cite{jiang2019black} are less effective, they utilize  the intrinsic movement pattern and regional relative motion, and propose the motion-aware noises to replace random noises.  By using this prior in gradient estimation, fewer queries are needed to perform video attacks.  Wei \textit{et al}. \cite{wei2019heuristic} search  {for} a subset of frames based on the importance of each video frame to the recognition model. Besides, they also limit the adversarial perturbations only on some salient regions. Because the temporal and spatial reductions are separately formulated, the method usually needs  hundreds of thousands of queries. To mitigate this defect.  Wei \textit{et al}. \cite{wei2022sparse} have proposed a sparse video attack algorithm based on reinforcement learning. An agent is designed to identify key frames through some interactions with the threat model. It can significantly reduce the adversarial perturbations, but update the agent only after each round of successful attack. This  poor update mechanism leads to  {many unnecessary queries} and a weak fooling rate.   {RLSB attack \cite{wang2021reinforcement} explores to select key frames and key regions to reduce the high computation cost. However, the reinforcement learning is only applied to select key frames, which is similar to \cite{wei2022sparse}. The process of selecting key regions is based on the saliency maps, it is independent to the process of selecting key frames, and not integrated into the reinforcement learning framework. Thus, the selecting key frames and key regions are separately formulated.} Recently, \cite{li2021adversarial} presents to parameterize the temporal structure of the search space using geometric transformations, and then reduce the temporal search space. Thus, they can efficiently estimate the gradients.

In this paper, we also explore important searching space, which is different from the previous work focusing only on  key frames in the temporal domain. We jointly consider the identification of key regions in the spatial domain besides the temporal domain. For that, a multi-agent reinforcement learning is designed to identify a reduced space through rewards on the inherent property of video and interactions with the threat model. The comparisons with query-based black-box video attack methods are summarized in Table \ref{tab:trans}.

\subsection{Spatial-Temporal Property for Videos}
Video can be regarded as multiple continuous images, therefore video processing often needs to  consider both spatial and temporal correlations. The simultaneous consideration of temporal and spatial correlation of video is the key of video related tasks. 
Video action recognition is a longstanding research topic in multimedia and computer vision. Many mainstream algorithms are motivated by the advances in image classification, and improved through utilizing the temporal dimension of the video data. To facilitate the classification performance, Wu \textit{et al}. \cite{wu2015modeling} have proposed a hybrid deep learning framework for video classification, which is able to harness not only the spatial and short-term motion features, but also the long-term temporal clues. They integrate the spatial and temporal features in deep neural model with elaborately designed regularizations to explore feature correlations. The method can produce competitive classification performance. Some  works based on the spatial-temporal property can be found in \cite{hara2018can, wang2016temporal, lin2019tsm}.

Unlike the above methods, we consider the spatial-temporal property of videos in the video attack task. The temporal and spatial redundancy within videos are reduced to improve the efficiency of video attack, which extends the application scope of spatial-temporal property of videos. 

\section{Methodology}\label{sec:method}

In this section, we first give the baseline video attack algorithm: PGD \cite{madry2017towards} attack with NES \cite{wierstra2008natural} gradient estimator. Then the details of integrating cooperative Multi-Agent Reinforcement Learning (MARL) \cite{lowe2017multi} into the baseline are introduced. Finally, the whole algorithm is summarized. 

\subsection{ Preliminaries}
We assume $F(\cdot)$ is a black-box video recognition model  only whose \emph{top-1} information including the category label and confidence score can be required. Given a video  $X=\{x_i|i=1,...,M\}$ with ground-truth label $y$ where $x_i\in\mathbb{R}^{H\times W\times 3}$ denotes the $i$-th frame, and $M$ is the total frame number, the predicted category label is $y=F(X)$, and the corresponding confidence score is $P(y|X)$.

To attack the video recognition model, we extend Projected Gradient Descent (PGD) \cite{madry2017towards} to adapt the video data. The adversarial video  $X'$  {under the un-targeted attack is defined as:
\begin{equation}
X'_{t+1}=Proj(X'_t+\alpha \cdot sign(\nabla_Xl(X'_t,y))),
\label{eq:pgd}
\end{equation}}
where $Proj(\cdot)$ projects the updated adversarial example to a valid range. $\alpha$ is the attack step, and  is used to control the magnitude of the added adversarial noise per each iteration.  The $sign(\cdot)$ is the sign function, and $l(\cdot)$ is the cross-entropy loss function. 
Due to the limitation of black-box settings, we cannot obtain the accurate gradient $g$ by directly computing $g=\nabla_Xl(X'_t,y))$. Instead, \cite{ilyas2018black} proposes to utilize Natural Evolution Strategy (NES) \cite{wierstra2008natural} to estimate $g$ by querying the threat model. Specifically, NES can be described as:
\begin{equation}
{g}\approx  \cfrac{1}{{\Delta n}} \sum_{i=1}^n \sigma_i \cdot P(y|X_t'+ \Delta \cdot \sigma_i).
\label{eq:nes}
\end{equation}
It first samples $n/2$ values $\delta_{i}\backsim N(0,I)$, and then sets $\delta_{j}=-\delta_{n-j+1}  , j\in\lbrace(n/2+1)  ,..n\rbrace$. Finally, the gradient ${g}$ is estimated through averaging the ratio of the predicted results to search variance $\Delta$.

 {For the targeted attack, Eq.(\ref{eq:pgd}) is modified as follows:
\begin{equation}
X'_{t+1}=Proj(X'_t-\alpha \cdot sign(\nabla_Xl(X'_t,y'))),
\label{eq:gradient2}
\end{equation}
where $y'$ is a target category label pred-defined by the adversary in advance. In Eq.(\ref{eq:nes}), the ground-truth $y$  should also be  modified as the target label $y'$ to estimate the gradients versus the target label.}

 {In practial application, directly performing Eq.(\ref{eq:nes}) is inefficient. Because the number of sample points $n$ is related with the dimension of $X_t'\in\mathbb{R}^{M\times H\times W\times 3}$. Owing to the high dimension of video data $X_t'$, we need to set a large value of $n$ to compute an accurate gradient in each iteration $t$, which will lead to a large number of queries with the threat model. To improve the attack efficiency, the video dimension should be reduced by selecting the key frames and key regions, obtaining a reduced $M, H$ and $W$, and thus a small value of $n$ can be available. Technically, we hope to replace $X_t'$ in Eq. (\ref{eq:nes}) with $\hat{X}_t'=\Gamma(X_t')$, where $\Gamma(\cdot)$ denotes the reduced operation, and $\hat{X}_t'$ is the reduced video.}
\begin{figure}[t]
\begin{center}
\includegraphics[width=0.9\linewidth]{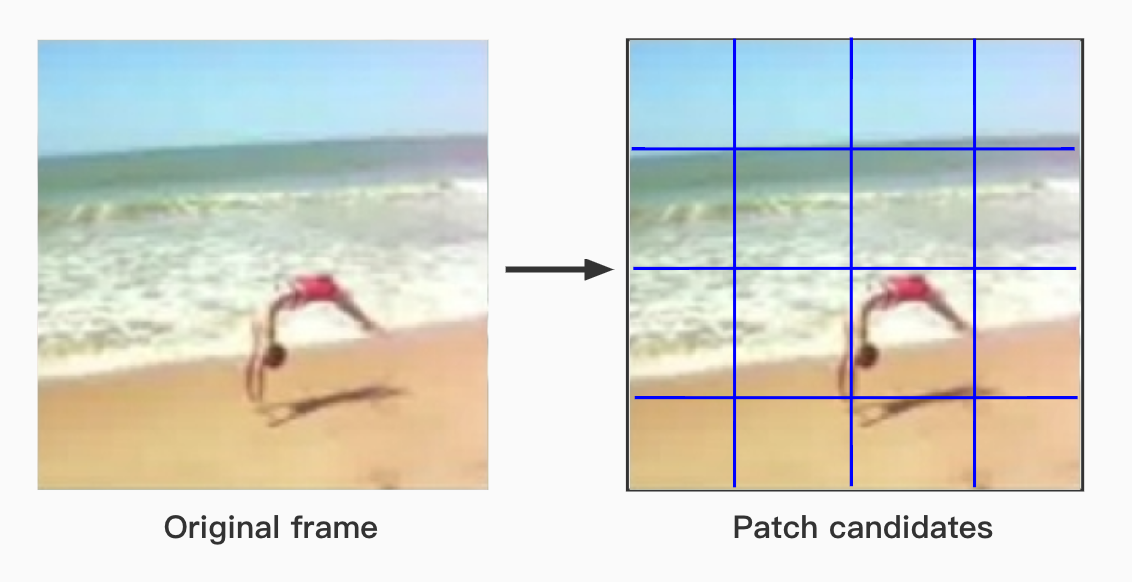}
\end{center}
\vspace{-0.5cm}
\caption{The designed actions of the spatial agent. In each frame, we uniformly divide the frame into overlapped patches according to a predefined stride. All the patch candidates constitute the actions. For simplicity, the stride equals to the patch size in this example. }
\label{fig:patch_action}
\vspace{-0.5cm}
\end{figure}

\subsection{The Proposed AstFocus Attack}
 {To implement the above idea, we build the so-called AstFocus attack based on a cooperative multi-agent reinforcement learning (MARL) to jointly solve for the key frames and key regions during the black-box attack process. In AstFocus attack,  one agent is responsible for selecting key frames (temporal agent), and another agent is responsible for selecting key regions (spatial agent). These two agents are cooperative to achieve the same goal.} The processes of selecting key frames and key regions in each iteration of PGD are formulated into the Markov Decision Processes (MDP). The details of these two agents as well as the optimization algorithm are given below.

\subsubsection{Spatial Agent}
 {Spatial agent actually aims at solving an object localization problem (detecting key regions)}, we detail from three parts. 

\textbf{Action Design:}
To construct the actions of the spatial agent, we uniformly divide each video frame into overlapped patches inspired by the Vision Transformer \cite{yuan2021tokens}. In this way, we obtain a candidate patch set for the $i$-th frame $x_i$: $B_i=\{b_i^j|j=1,...,D\}$ where $b_i^j$ denotes the $j$-th patch region within $x_i$, and $D$ is the total number of candidate patches in this frame. $b_i^j\in\mathbb{R}^{h\times w}$ denotes that the patch's size is $h$ and $w$, and their values will be tuned in the experiments. The goal of spatial agent is to select an optimal patch $b_i^{*} \in B_i$ in each frame as the key region, and thus the final selected action is a sequence set $a^p=\{b_i^{*}|i=1,...,M\}$. From the definition, we can see that there are totally $D^M$ action combinations for the given video $X$, which implies the search space is huge. An example of actions in one frame is listed in Figure \ref{fig:patch_action}, where $D=16$. 

\begin{figure}[t]
\centering
\includegraphics[width=0.95\linewidth]{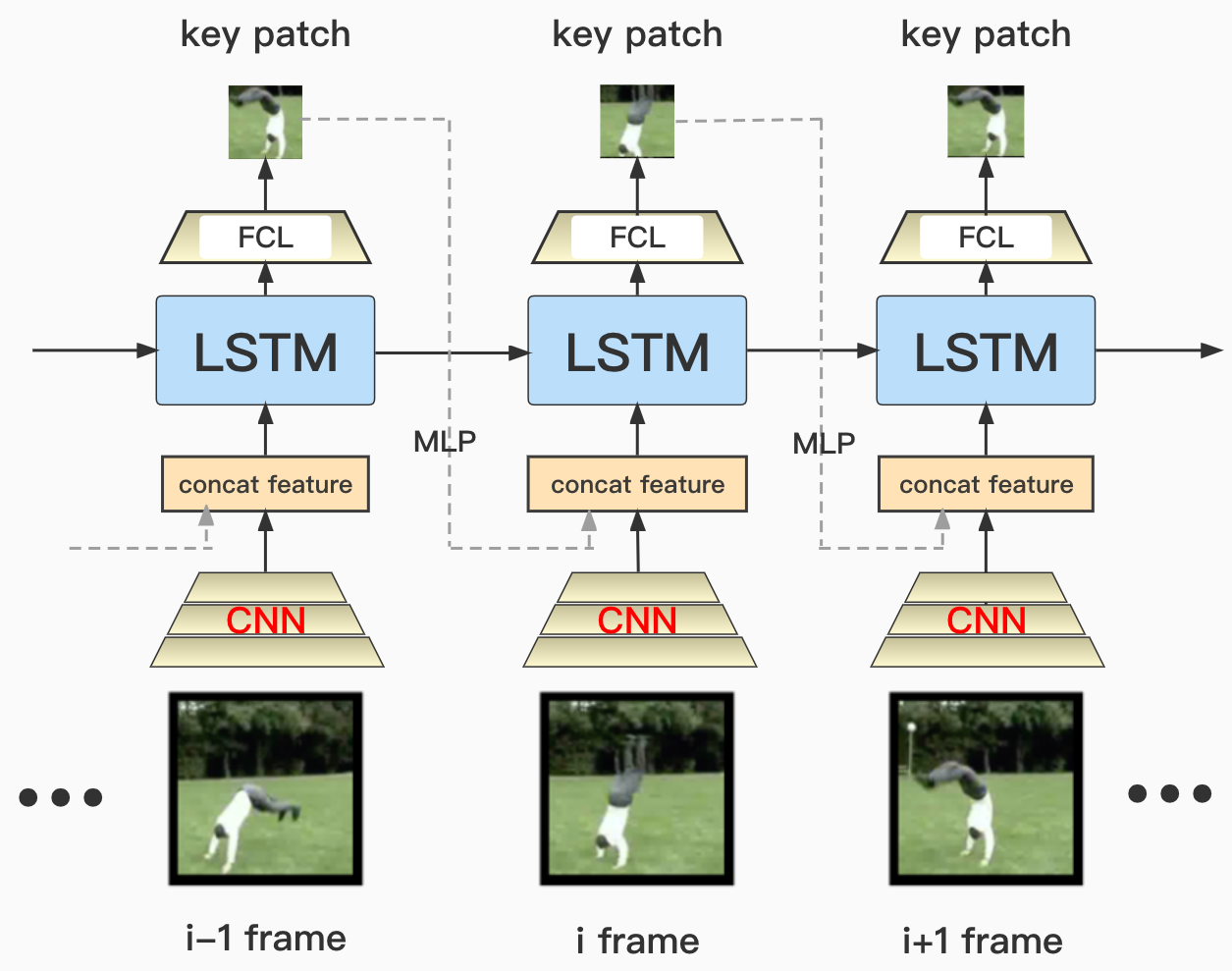}
\vspace{-0.5cm}
\caption{The flowchart for the Policy network of the proposed spatial agent. It is used to identify the crucial regions of each video frame.}
\label{fig:patch select}
\vspace{-0.5cm}
\end{figure}

\textbf{Policy Network Design:} 
Spatial policy network $\pi^p(a^p|s^p)$ is used to predict the spatial action $a^p$ when the state $s^p$ is given. The flowchart of our policy network is listed in Figure \ref{fig:patch select}. Overall speaking, because we need to tackle with the sequence video data, a LSTM-based \cite{greff2016lstm} structure is used to construct the policy network $\pi^p(a^p|s^p)$. For the $i$-th frame $x_i$, a lightweight convolution neural network (CNN) $f(x_i)$ is first to extract the frame-level feature maps $e_i$. In our experiments, we use MobileNet V2 as the lightweight CNN backbone for simplicity. Users can also apply other lightweight CNNs. Then they are fed into the LSTM unit to predict the logits for each patch. Next, a Softmax with Fully Connected Layer (FCL) is attached to output each patch's probability $p_{b_i^j}$. Finally, we utilize the categorical sampling to obtain the optimal patch region $b_i^{*}$ according to their probability values $p(B_i)=\{p_{b_i^j}|j=1,...,D\}$. To guarantee the smooth change of selected patch between adjacent frames, we concat the local patch features $e^{b^*}_{i-1}$ of the previous selected patch $b_{i-1}^*$ with the current frame-level features $e_i$ to jointly predict the current patch region, and $e^*_{i-1}$ is extracted via a simple multilayer perceptron (MLP) on the corresponding patch features of $e_{i-1}$. 

Formally, the frame-level feature maps are extracted by:
\begin{equation}
e_i = f(x_i), i=1,2,...,M,
\label{eq:backbone}
\end{equation}
next, the optimal action for each frame is achieved by:
\begin{equation}
p(B_i) = \pi^p_{\theta}(\cdot|concat(e_i,e^{b^*}_{i-1}),h^{\pi}_{i-1}), i=1,2,...,M,
\label{eq:policy1}
\end{equation}
\begin{equation}
b_i^{*} = categorical(p(B_i)), i=1,2,...,M,
\label{eq:policy}
\end{equation}
where $h^{\pi}_{i-1}$ denotes hidden states output by LSTM unit in the $i$-1-th frame. Thus, the state $s^p$ in our method is defined as the concat feature $concat(e_i,e^{b^*}_{i-1})$. Eq.(\ref{eq:policy}) is repeated $M$ times to achieve the optimal action $a^p=\{b_i^{*}|i=1,...,M\}$. 

In our method, the policy network is updated in each iteration $t$ of PGD attack, therefore, the optimal action will be updated in each iteration until the PGD attack stops. 

\textbf{Reward Design:}
In each iteration, the spatial policy network will receive the feedback from the environment to update its parameters $\theta^p$. Therefore, we need to design the reasonable rewards to guide the update of policy network. Because AstFocus attacks are based on the Multi-Agent Reinforcement Learning (MARL) framework, we design two kinds of rewards: one is specific for the spatial agent, and another is the common rewards shared with temporal agent. 

For the special reward, an intuitive idea to evaluate the patch's importance is the area covered by the foreground objects. Because video recognition model mainly performs predictions based on the foreground objects like person, car, etc. Therefore, if the policy network $\pi^p(a^p|s^p)$ selects the foreground patch, the specific reward should be enlarged, and thus the policy network will be encouraged to select the foreground object in the next iteration. Based on this idea, we need a metric to measure the objectness score for a given patch. We here choose a classic objectness model: edgeboxes \cite{zitnick2014edge}. It calculates the edge response of each pixel and determines the boundary of the object by using the structured edge detector. 

More concretely, the $r^i_{edgebox}$ reward for the selected patch $b_i^*$ can be described as following:
\begin{equation}
r^i_{edgebox} =  \cfrac{\prod_{k}w_b(s_k)\cdot u_k}{ 2\cdot(w+h)^2}, i=1,2,...,M,
\end{equation}
The edgebox reward for the whole video is defined:
\begin{equation}
r_{edgebox} = \sum_{i=1}^M r^i_{edgebox}, i=1,2,...,M,
\label{eq:edgebox}
\end{equation}
where $w $ and $h $ are the patch’s width and height. $w_b(s_k)$ is used to measure the affinity of the $k$-th edge groups in the selected patch. The $u_k$ is the sum of the $k$-th edge groups in the selected patch. In general, a large patch often results in a large edgebox value. More detailed information about edgebox function can be found in \cite{zitnick2014edge}.

For the common reward, it comes from the feedback of the black-box threat models. If the selected patch is reasonable, the generated adversarial patch should have a strong attacking ability, and thus will make the confidence score output by threat models have a big drop. Therefore, we can use the confidence drop of the ground-truth label as a metric to compute this reward. Because this is also useful to the temporal agent, it is called as common reward. Specifically, the common reward $r_{common}$ is defined as follows:
\begin{equation}
V(X')=exp(P(y'|X')-P(y|X'));
\label{eq:common1}
\end{equation}
\begin{equation}
r_{common} = \cfrac{V(X'_{t+1})-V(X'_{t})}{V(X'_t)},
\label{eq:common}
\end{equation}
where $exp(\cdot)$ is the exponential function, and $P(y|X')$ represents the ground-truth label's confidence when $X'$ is fed into video recognition model.  {In the un-targeted attack, $P(y'|X')$ represents the second-ranked label's confidence which is considered as the most competitive label to replace the ground-truth label. Only if the second-ranked label's confidence becomes larger than that of the ground-truth label, $V(X')$ becomes large}. Then, we use the relative change of $V(X')$ at different iteration $t$ as the metric for common reward. Eq.(\ref{eq:common}) is designed to encourage the agent to add perturbations on the selected regions that make the second-ranked label's confidence gradually reach the ground-truth label and finally exceed it.  {In targeted attack, $P(y'|X')$ is the confidence of pre-defined target label by the adversary.} 

In summary, the $t$-iteration reward for spatial agent is:
\begin{equation}
r^t_{spatial} = r^t_{common} + \lambda_1r^t_{edgebox},
\label{eq:spatial_reward}
\end{equation}
where $\lambda_1$ denotes a weight to balance the two terms. 

\subsubsection{Temporal Agent}
 {There exists a major distinction between temporal agent and spatial agent. Spatial agent aims at solving the object localization problem, while temporal agent aims at solving the binary classification problem (selecting or not selecting a frame). Thus, the actions, rewards, and policy networks of temporal agent should be re-designed.}

\textbf{Action Design:}
Key frames refer to those video frames that are conducive to successful attack, and their number is less than that of the whole video. The goal of the temporal agent is to select some key frames from the whole input video $X$, and thus the final selected action is also a sequence set $a^f=\{o_i^*|i=1,...,M\}$ just like spatial agent. The $o_i^* \in \{0, 1\}$ indicates whether the $i$-th frame is selected or not. Therefore, there are totally $2^M$ different actions, which is not friendly to direct optimization learning.
\begin{figure}[t]
\begin{center}
\includegraphics[width=0.95\linewidth]{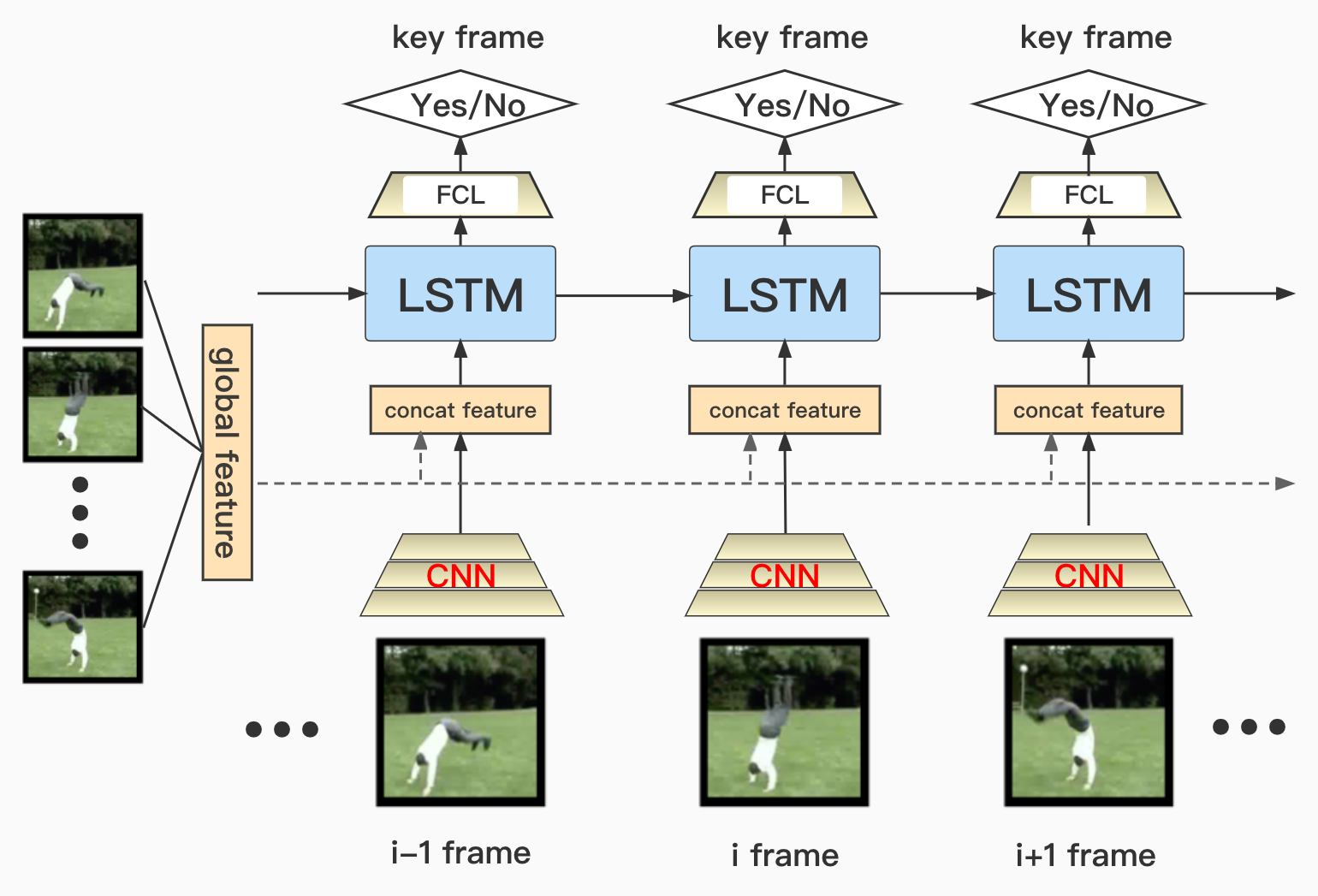}
\end{center}
\vspace{-0.5cm}
\caption{The flowchart for the policy network of the proposed temporal agent. It is used to select the key video frames from the input video.}
\label{fig:frame_selection}
\vspace{-0.2cm}
\end{figure}

\textbf{Policy Network Design:} 
The temporal policy network $\pi^f(a^f|s^f)$ is used to predict the spatial action $a^f$ when the state $s^f$ is given. It is constructed with a LSTM structure. The skeleton diagram of the temporal policy network is shown in Figure. \ref{fig:frame_selection}. The input of the policy network is the concat features composed of current frame-level features $e_i$ and a video-level global features $e^g$. Combining these two features can better select the key frames by considering the global video information. The global features $e^g$ is achieved by a fully connected layer on all the frame-level features $e_i$, $i=1,...,M$. The output of LSTM network is then fed to a Softmax with Fully Connected Layer (FCL) to predict the probability $p_i$ to indicate $o_i$=1.
Technically, the temporal policy network can be expressed as:
\begin{equation}
p_i = \pi^f_{\theta}(\cdot|concat(e_i,e^g),h^{\pi}_{i-1}), i=1,2,...,M,
\label{eq:tempolicy}
\end{equation}
\begin{equation}
o_i^* = Bernoulli(p_i), i=1,2,...,M,
\label{eq:bernouli}
\end{equation}
where $Bernoulli(\cdot)$ is the Bernouli function. $h^{\pi}_{i-1}$ denotes the hidden states output by LSTM unit in the $i$-1-th frame. The state $s^f$ is defined as the concat feature $concat(e_i,e^g)$. Eq.(\ref{eq:bernouli}) is repeated $M$ times to get  $a^f=\{o_i^{*}|i=1,...,M\}$.

\textbf{Reward Design:}
To make the temporal agent intelligent, the temporal policy network interact with the environment for updating its parameters $\theta^f$.  Similar to the training of the spatial agent, in addition to the common reward function which shared with the spatial agent, we also have designed two special rewards to guide the temporal agent. The first specific reward function is the sparse reward $r_{sparse}$:
\begin{equation}
r_{sparse} = exp(- \dfrac{1}{M}|\sum_{i=1}^{M} o_{i} - L |),
\label{eq:upper_bound}
\end{equation}
where $L$ here is used to control the number of key frames selected by the temporal agent, and $L < M$.  The second specially designed reward function is mainly used to evaluate the representative ability of the video frames selected by the temporal agent. Because the selected video frames need to be sparse but effectively represent the semantic information of the whole input video. The representative reward function \cite{zhou2018deep} $r_{rep}$ is defined as:
\begin{equation}
r_{rep} = exp(-\frac {1}{M}\sum_{i=1}^M \min\limits_{t' \subset \mathcal{K} } \vert\vert e_i - e_{t'} \vert\vert_2),
\label{eq:rep}
\end{equation}
where $\mathcal{K}$ is a set of selected frame, i.e.,frames with $o_i$=1. Through those reward functions, the temporal can be forced to recognize few and critical video frames. The selected video frames can be effectively reduce the temporal redundancy of the entire video and effectively improve the following attack efficiency.

To make key video frames conducive to successful attacks, $r_{common}$ also join the learning of the temporal agent. For the $t$-th iteration, the corresponding reward is:
\begin{equation}
r^t_{temporal} = r^t_{common} + \lambda_2r^t_{sparse} + \lambda_3r^t_{rep},
\label{eq:temporal_reward}
\end{equation}
where $\lambda_2$ and $\lambda_3$ are two balance coefficients, and they will be discussed and set up in the experimental section.

So far, through the cooperation of spatial agent and temporal agent, the key regions in the key video frames of the input video can be identified. In the procedure of multi-agent reinforcement learning, the agents interact with the threat model for many times, and the predicted results of the agents are more inclined to the rapid and successful attack. Therefore, the critical attacking spaces selected by multi-agent reinforcement learning is the space sensitive to attack, which can effectively improve the efficiency of attack.

\subsubsection{Optimization Algorithm}
There are two parts to optimize: one is the lightweight CNN backbone $f(\cdot)$, and another is the policy network $\pi(\cdot)$. 

\textbf{CNN backbone:}
In our method, the CNN backbone $f(\cdot)$ is used for both temporal agent and spatial agent. It extracts the frames' feature maps to construct  state $s$. To decouple the training process of CNN backbone and policy network, we directly apply a pre-trained MobileNet V2 backbone on ImageNet dataset as the feature extractor. In this way, we can focus on the optimization of two policy networks. 

\textbf{Policy network:}
The policy gradient methods are used to optimize the temporal and spatial policy network. They are to directly adjust the parameters $\theta$ in order to maximize the objective $J(\theta)= \mathbb{E}_{s\backsim \rho^{\pi},a\backsim \pi_{\theta}}[R]$ by tacking steps in the direction of $\nabla_{\theta}J(\theta)$. By introducing an  action-value function $Q^{\pi}(s,a)$, the policy gradient can be changed as:
\begin{equation}
  \nabla_{\theta}J(\theta)=  \mathbb{E}_{s\backsim \rho^{\pi},a\backsim \pi_{\theta}}[\nabla_{\theta}log\pi_{\theta}(a|s)Q^{\pi}(s,a)].
  \label{eq:pga}
\end{equation}
To solve Eq.(\ref{eq:pga}), we utilize the actor-critic reinforcement learning framework \cite{konda1999actor} where a critic network is applied to approximate the action-value function $Q^{\pi}(s,a)$.  Actor network is the policy network in Figure \ref{fig:patch select} and Figure \ref{fig:frame_selection}. 

Because our method focuses on the cooperative multi-agent tasks, in which the two agents are trying to optimize a shared reward function. Each agent is decentralized and only has access to locally available information. For example, temporal agent can only observe the change of key frames, and spatial agent can only observe the change of key patches. Therefore, our method can be described as Decentralized Partially Observable Markov Decision Processes (Dec-POMDP) \cite{spaan2012partially}. To solve this problem, \cite{lowe2017multi} presents the multi-agent decentralized actor, centralized critic approach, thus Eq.(\ref{eq:pga}) is reformulated as:
\begin{equation}
  \nabla_{\theta_k}J(\theta_k)=  \mathbb{E}_{s\backsim \rho^{\pi},a_k\backsim \pi_{k}}[\nabla_{\theta_k}log\pi_{k}(a_k|s_k)Q_k^{\pi}(\textbf{s},a_1,...,a_K)].
  \label{eq:mapga}
\end{equation}
where $\pi_k$ denotes the policy network of the $k$-th agent, and $\theta_k$ is the corresponding parameters. In our method, there are totally two agents. The corresponding policy networks are $\pi^p(a^p|s^p)$ with parameter $\theta^p$ and   $\pi^f(a^f|s^f)$ with parameter $\theta^f$.  {Here $Q_k^{\pi}(\textbf{s},a_1, a_2)]$ is a centralized action-value function that takes as input the actions of all agents $(a_1,a_2)$ in addition to some state information $\textbf{s}=[s^p,s^f]$, and outputs the Q-value for agent $k$. In this way, we can perform a communication between two agents. In cooperative MARL, each agent is expected to maximize the common reward and its specific reward, therefore, we just need to solve Eq.(\ref{eq:mapga}) according to the rewards for spatial agent and temporal agent, respectively}.

\begin{algorithm}[t]
	\renewcommand{\algorithmicrequire}{\textbf{Input:}}
	\renewcommand{\algorithmicensure}{\textbf{Output:}}
	\caption{AstFocus black-box video attack algorithm}
	\label{alg::conjugateGradient}
	\begin{algorithmic}[1]
		\REQUIRE{Clean video:$X$; ground-truth label:$y$; feature extractor: $f(\cdot)$;
      black-box video recognition model:$F(\cdot)$.
      Max PGD iterations:$T$; PGD attack step:$\alpha$;
      learning rate:$\epsilon$.}
		\ENSURE Adversarial video $X'$
		\STATE {Initialize parameters $\theta^f$ and $\theta^p$ for temporal policy network $\pi^f(\cdot)$ and spatial policy network $\pi^p(\cdot)$};
        \STATE{Extract frame-level features $\{e_i|i$=1,...,$M\}$ via Eq.(\ref{eq:backbone});}  
		\WHILE{$t < T$}
                 \STATE{Compute key regions $a^p=\{b_i^{*}|i=1,...,M\}$ via Eq.(\ref{eq:policy1}) and Eq.(\ref{eq:policy});}
                 \STATE{Compute key frames $a^f=\{o_i^{*}|i=1,...,M\}$ via Eq.(\ref{eq:tempolicy}) and Eq.(\ref{eq:bernouli});}
                 \STATE{Obtain the core video$\hat{X}=\{o_i^{*}\cdot b_i^{*}|i=1,...,M\}$;}
                 \STATE{Estimate the gradients on $\hat{X}$ via Eq.(\ref{eq:nes});}
                 \STATE{Generate adversarial video $X'_{t+1}$ via Eq.(\ref{eq:pgd});}
            \IF{$F(X'_{t+1})\neq y$ }
            \STATE{$X' \leftarrow X'_{t+1}$; Break;}
            \ELSE
            \STATE{Compute spatial reward $r^t_{spatial}$ via Eq.(\ref{eq:spatial_reward}) and temporal reward $r^t_{temporal}$ via Eq.(\ref{eq:temporal_reward});}            
            \STATE{Compute $\nabla_{\theta^f}J(\theta^f)$ and $\nabla_{\theta^p}J(\theta^p)$ via Eq.(\ref{eq:mapga});}
            \STATE{Update $\theta^f \leftarrow \theta^f$ + $\epsilon\cdot \nabla_{\theta^f}J(\theta^f$);}
            \STATE{Update $\theta^p \leftarrow \theta^p$ + $\epsilon\cdot \nabla_{\theta^p}J(\theta^p$);}
            \ENDIF
	    \ENDWHILE
		\RETURN $X'$
	\end{algorithmic}  
\end{algorithm}

To solve Eq. (\ref{eq:mapga}) for the spatial agent $\pi^p(a^p|s^p)$ and temporal agent $\pi^f(a^f|s^f)$, we use the Proximal Policy Optimization (PPO), a popular single-agent on-policy RL algorithm \cite{schulman2017proximal} to obtain the $\theta_f$ and $\theta_p$. For the details of PPO algorithms, please refer to \cite{schulman2017proximal}. 

\subsection{The Overall Framework}
After the MARL module, the key frames and key regions are obtained.  {The  video $X_t'\in\mathbb{R}^{M\times H\times W\times 3}$ in Eq.(\ref{eq:pgd}) and Eq.(\ref{eq:nes}) is reduced to the video $\hat{X}_t'\in\mathbb{R}^{m\times h\times w\times 3}$ composed of key frames and key regions, where $m$ denotes the number of key frames, and $h, w$ denote the key patches' height and width.} It is clear that $m \ll M, h \ll H, w \ll W$. AstFocus attack finally utilizes $\hat{X}_t'\in\mathbb{R}^{m\times h\times w\times 3}$ to compute Eq.(\ref{eq:nes}). Because of the reduced dimension, the gradient estimation can be efficient.  

We now give the overall algorithm of AstFocus attack, which is illustrated under the un-targeted attack. The process of agent learning is an unsupervised process. Through continuous interaction with the threat model, the agent gets the feedback from the attack effect and external evaluation indicators from the video itself to update agents to encourage them to perform better. The whole algorithm is summarized in Algorithm \ref{alg::conjugateGradient}.

\section{Experiments and Results}

\subsection{Datasets and Recognition Models}

\textbf{Datasets.}  {In our experiments, three public action recognition datasets: UCF-101 \cite{soomro2012ucf101}, HMDB-51 \cite{kuehne2011hmdb}, and Kinetics-400 \cite{carreira2017quo} are used. The UCF-101 contains 13,320 videos with 101 action categories, HMDB-51 is a dataset for human motion recognition, which contains 51 action categories with a total of 70,00 videos. Kinetics-400 contains 400 human action classes, with at least 400 video clips for each action. All of these datasets divides 70\% of the video into training sets and 30\% of the test sets. We randomly sample 100 videos from UCF-101 test set, 50 videos from HMDB-51 test set, and 400 videos from Kinetics-400 test set. All sampled videos can be classified by the recognition models correctly.}

\textbf{Recognition Models.}  {For recognition models, four representative methods are used in our experiments. They are C3D \cite{hara2018can}, Temporal Segment Network (TSN) \cite{wang2016temporal}, Temporal Shift Module (TSM) \cite{lin2019tsm}, and SlowFast network \cite{feichtenhofer2019slowfast}. These models are all mainstream methods for video classification task. For TSN, TSM, and SlowFast on three datasets, we utilize the corresponding pre-trained  weights released by MMAction2 \cite{2020mmaction2}, a widely used open-source toolbox for video understanding based on PyTorch. For C3D, because MMAction2 only releases the pre-trained weights on UCF101, to ensure the consistency, we utilize the officially pre-trained weights on three datasets released by the authors\footnote{https://github.com/kenshohara/3D-ResNets-PyTorch}.  Table \ref{accuracy-test} lists their accuracy values under the test set.}

\begin{table}[t]
\caption{ {The accuracy of four different modes on three datasets.} }
\center 
\begin{tabular}{cccc}
\hline
\multirow{2}{*}{Models} & \multicolumn{3}{c}{Datasets} \\ \cline{2-4} 
 & UCF-101 & HMDB-51  &Kinetics-400\\ \hline
C3D \cite{hara2018can} &  85.88\% &  59.57\%  &54.20\%\\ \hline
TSN \cite{wang2016temporal} &  83.03\% &56.08  \%   &70.42\% \\ \hline
TSM \cite{lin2019tsm} &  94.58\% &  74.77\%   &71.90\% \\ \hline
SlowFast \cite{feichtenhofer2019slowfast} &  92.78\% &  65.95\%   &74.42\% \\ \hline
\end{tabular}
\label{accuracy-test}
\end{table}

\subsection{Evaluation metrics}
There are  {four} metrics to test the performance of our method on various sides. Specifically, Fooling Rate, Query Number, Mean Absolute Perturbation, and Time are explored.

\textbf{Fooling Rate (FR):}  indicates the percentage of adversarial videos, which successfully fool the threat model, out of all the tested videos. FR reflects the probability of successfully generating adversarial  examples.   A higher FR value means the better performance on the task of attacks.

\textbf{Mean Absolute Perturbation (MAP):} denotes the  magnitude value of the generated adversarial perturbation $\mathbf{r}$. For a given video: MAP=$\frac{1}{M}\sum_i|\mathbf{r}_i|$, where $M$ is the number of frames in a video, and $\mathbf{r}_i$ is the perturbation intensity vector on the $i$-th frame.  To be intuitive, the value of MAP is resized to 0-255. In the experiments, we report the average MAP across the test videos.  A lower MAP value means the better imperceptibility.

\textbf{Query Number (QN):} denotes the used query times to successfully fool the threat model for a given adversarial video. It reflects the efficiency of different video attack methods. In the experiments, we set an upper bound for the query number, if the queries reach the upper bound but the threat video model is still not fooled successfully, we think this adversarial video is not successfully generated. The average query number across the test videos is reported.  A lower QN value means the higher efficiency.

 {\textbf{Time (T):} denotes all the cost time when the successful attack is finished. We use seconds to measure the time.  In the experiments, we report the average seconds across the test videos.  A lower time value means the higher efficiency.}

Note that previous works \cite{yan2021efficient,wei2022sparse, wei2019heuristic} have also used these metrics. But this paper has a slight difference with them. In \cite{yan2021efficient,wei2022sparse,wei2019heuristic}, they compute MAP and NQ values only for the adversarial videos which can successfully perform the attacks, the  {MAP and NQ values of} failed videos  don't be considered. In contrast, this paper computes MAP and NQ values for all the test videos. We think this is more reasonable because the failed videos also generate perturbations and cost queries with the threat models. 

\subsection{State-of-the-art attack competitors}
Here, we use  {six} state-of-the-art black-box video attack methods as comparisons with our method in effect and speed, named VBAD attack \cite{jiang2019black}, Heuristic attack \cite{wei2019heuristic}, Sparse attack \cite{wei2022sparse}, GEO-TRAP attack \cite{li2021adversarial},  {RLSB attack \cite{wang2021reinforcement}}, and Motion-sampler attack \cite{zhang2020motion}. The detailed introductions about these competitors can be found in the related works section. We use their own officially released codes to conduct comparisons (for Sparse attack, we directly use the well-trained agent to predict key frames and then perform attacks. There is no released code  for RLSB attack, we implement it according to the paper). For fair comparisons, all the settings are the same. 

\subsection{Implementation details} 
In the query-based black-box attacks, the query number is a key metric to evaluate the attacks' performance. Thus,  given a video, we set a maximum query number for all the compared methods. If the used query number is above the maximum query number, the adversarial attack is regarded as failure for this given video. We here set the maximum query number  to $1.5\times 10^4$ in the un-targeted attack and $3\times 10^4$ in the targeted attack. In the NES, we set the variance $\Delta $ in NES to $10^{-3}$ for the un-targeted attack and $10^{-6}$ for the targeted attack according to our experience.

\subsection{Parameter tuning} 
There are some hyperparameters in our method. In this section, we will determine their values via parameter tuning on the validate set. Specifically, we randomly selected 20 videos from HMDB-51 to construct the validation set, and then perform parameter tuning versus C3D  model. 
\begin{figure}[t]
\center
\includegraphics[width=1\linewidth]{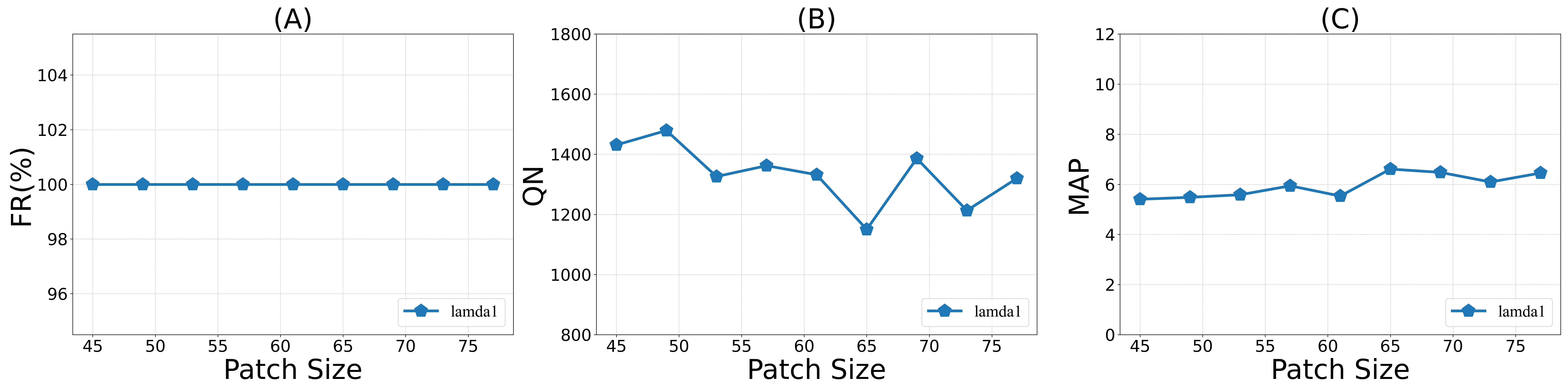}
\caption{The parameter tuning results of AstFocus attacks with different patch sizes. (A) The effects for fooling rate. (B) The effects for query number. (C) The effects for perturbation magnitude. }
\label{fig:Patch_Size}
\end{figure}
\subsubsection{Patch size for spatial agent}
The first hyperparameter is the patch size $h$ and $w$ when designing the spatial agent's action. A reasonable patch size will lead to less queries and less perturbations. The parameter tuning results for patch size is given in Figure \ref{fig:Patch_Size}, where we explore its effects for the fooling rate, query number and perturbations, respectively. From the figure, we see that patch size mainly affects the query number but shows slight changes for fooling rate and perturbation magnitude.  {Moreover, Figure \ref{fig:Patch_Size} (B) shows the query number is relatively sensitive to the patch size. This is reasonable because the pre-defined patch size determines the proportion of selected key regions out of the whole image, thus affects the query number.}\footnote{ {From the 
last column in Table \ref{ta:ablation_results}, we see that AstFocus attack has smaller variance when performing multiple times. For attacking C3D model on HMDB-51, the variance has no changes for FR, and only changes 1\% around the mean for MAP,  4\% around the mean for NQ. Therefore, the unsmooth curve is not caused by the significant variance.}}  Overall, when the patch size is set to 65, the query number reaches the smallest value. Therefore, we set $h=w=65$.

\subsubsection{Upper bound of key frames}
\begin{figure}[t]
\center
\includegraphics[width=1\linewidth]{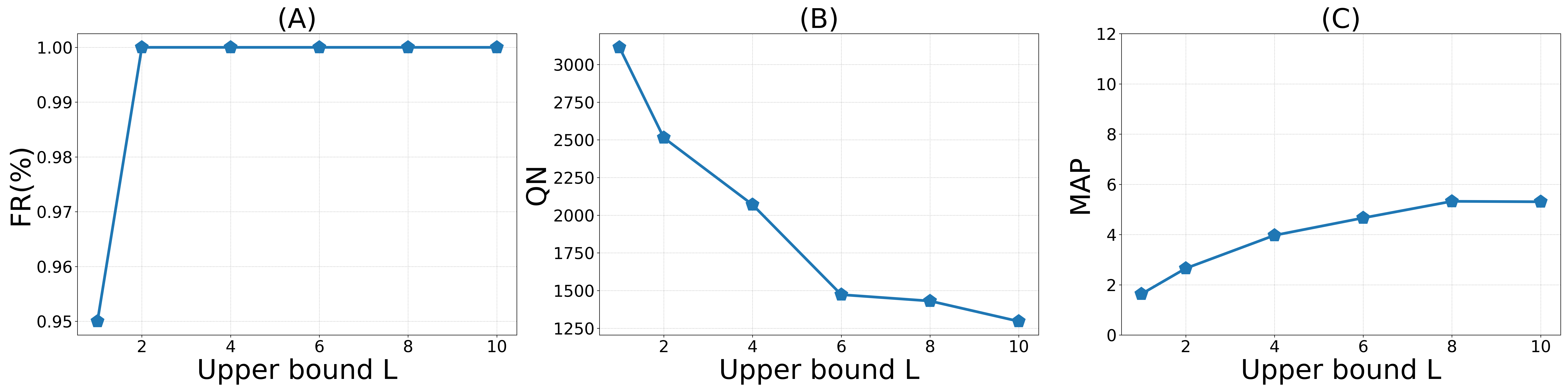}
\caption{The parameter tuning results of AstFocus attacks with different upper bounds of key frames. (A) The effects for fooling rate. (B) The effects for query number. (C) The effects for perturbation magnitude. }
\label{fig:K}
\end{figure}

The second hyperparameter is the upper bound $L$ of selected key frames in Eq.(\ref{eq:upper_bound}). A reasonable $L$ can help our method select the minimal key frames to perform a successful video attack, and thus query number can be reduced.  The parameter tuning results for upper bound $L$ is given in Figure \ref{fig:K}, where we also explore its effects for the fooling rate, query number and perturbations, respectively. We can see that with the increase of $L$ value, the fooling rate  will gradually stabilize to 100\% and query number is slowly decreasing, but it would cause a big increase in perturbation magnitude.  To  balance three different evaluation metrics, we set $L=10$ in the following experiments. 

\subsubsection{Sample number in NES}
\begin{figure}[t]
\center
\includegraphics[width=1\linewidth]{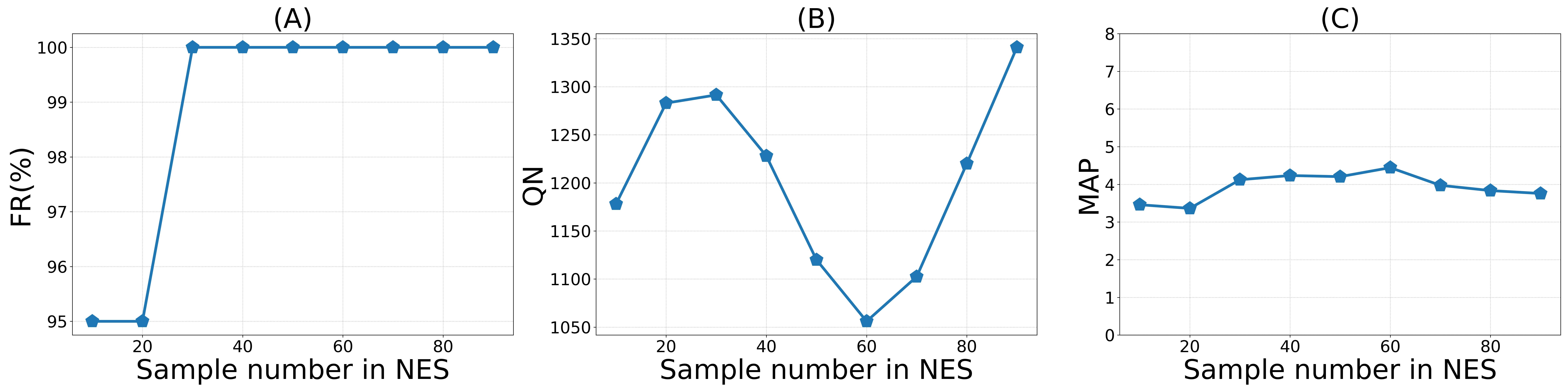}
\caption{The parameter tuning results of AstFocus attacks with different sample numbers $n$. (A) The effects for fooling rate. (B) The effects for query number. (C) The effects for perturbation magnitude. }
\label{fig:F_sample}
\end{figure}

The third hyperparameter is the number of sampled points $n$ in Eq.(\ref{eq:nes}). The sample number $n$ per each iteration has a great influence on the accuracy of the estimated gradient, especially when the attacking space changes. To explore the impact of the sample number $n$ on the attack effect, we have conducted a series of experiments. The parameter tuning results are given in Figure \ref{fig:F_sample}. We can see that with the increase of $n$ value, the fooling rate will gradually stabilize to 100\%,  but query number and perturbation magnitude achieve their optimal performance when $n$ is located in 60.  Therefore, we set $n=60$ in the following experiments.

\subsubsection{Weights for various rewards}
\begin{figure}[t]
\center
\includegraphics[width=1\linewidth]{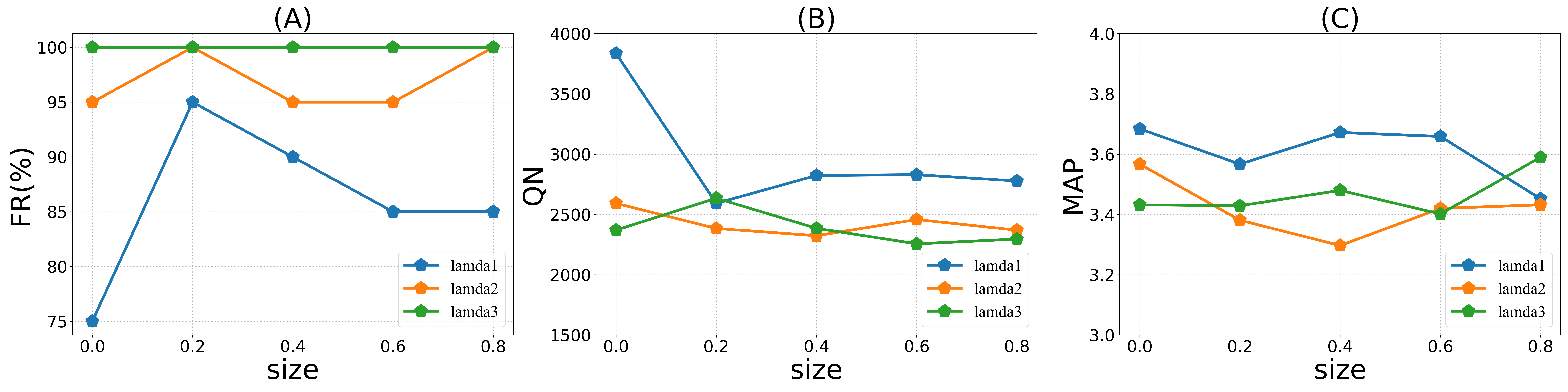}
\caption{The parameter tuning results of AstFocus attacks with different reward weights. (A) The effects for fooling rate. (B) The effects for query numbers. (C) The effects for perturbation magnitude. }
\label{fig:lamda}
\end{figure}

There are three weights to tune in the reward functions. They are 
$\lambda_1$ in Eq.(\ref{eq:spatial_reward}), $\lambda_2$ and $\lambda_3$ in Eq.(\ref{eq:temporal_reward}), which measures the importance of their own rewards. The  parameter tuning results are given in Figure \ref{fig:lamda}. According to the figure, we set $\lambda_1=0.2, \lambda_2=0.4$, and $\lambda_3=0.6$, respectively.  {It means there exists more redundancy to reduce in the temporal domain than spatial domain, thus needing to set large rewards in Eq.(\ref{eq:temporal_reward}) to guide agent for  learning  key frames. }

\subsection{Ablation study}
 To explore the effectiveness of different components in the proposed algorithm, a series of  experiments are conducted here. Specifically, we investigate the effects of various agents and various rewards, respectively. Similarly, we randomly select 20 videos from HMDB-51 to construct the validation set, and then perform the ablation study versus C3D video recognition model.  {Because the gradient estimator module introduces randomness, to consider this factor, we perform 
 each ablation study for five times, and then report the mean $\pm$ variance for different metrics.}

\begin{table}[t]
\caption{ {Effects of various agents to AstFocus attacks in an un-targeted setting.} }
\vspace{-0.5cm}
\center 
\setlength{\tabcolsep}{1 mm}{
\begin{tabular}{ccccc}
\hline
\multirow{2}{*}{Metrics} & \multicolumn{4}{|c}{Different agent versions}                                                                   \\ \cline{2-5} 
                         & \multicolumn{1}{|c|}{Baseline}   & \multicolumn{1}{c|}{Spatial}    & \multicolumn{1}{c|}{Temporal}   &Spatial\&Temporal   \\ \hline
FR(\%)                    & \multicolumn{1}{|c|}{ 73.3$\pm$2.9}   & \multicolumn{1}{c|}{88.3$\pm$2.9}    & \multicolumn{1}{c|}{93.3$\pm$2.9}   & 100$\pm$0.0    \\ \hline
QN                        & \multicolumn{1}{|c|}{3662$\pm$244}  & \multicolumn{1}{c|}{2670$\pm$155} & \multicolumn{1}{c|}{2757$\pm$125}  & 2227$\pm$92 \\ \hline
MAP                      & \multicolumn{1}{|c|}{6.37$\pm$0.08} & \multicolumn{1}{c|}{4.21$\pm$0.05}  & \multicolumn{1}{c|}{4.53$\pm$0.07} &3.35$\pm$0.03 \\ \hline
\end{tabular}}
\label{ta:ablation_results}
\end{table}

\subsubsection{Effects of various agents}
In our method, the baseline is PGD+NES algorithm. Then we integrate two agents into the PGD+NES to reduce the video dimension from the temporal and spatial domains, respectively. Here we perform the ablation study about  whether these two agents work in the video attack. The results are given in Table \ref{ta:ablation_results}, where ``Baseline" denotes the PGD+NES.  {In this setting, because there is no dimension reduction module, the perturbations are added on the whole video, which can be called as ``dense attack".} The term ``Spatial" denotes integrating the spatial agent into the baseline. In this setting, we reduce the spatial redundancy by selecting the key patches in each frame. The term ``Temporal" denotes integrating the temporal agent into the baseline, which reduces the temporal redundancy by selecting the key frames. The term ``Spatial$\&$Temporal" denotes the full version of AstFocus attack, i.e., simultaneously reducing the temporal and spatial redundancy via two agents.  

We show the effects versus fooling rate (FR), number query (NQ), and perturbation magnitude (MAP). From the table, we can see that the dimension reduction is indeed useful to the attacking performance, i.e., the ``Spatial" and ``Temporal" achieve higher FR and smaller QN and MAP than the ``Baseline".  By simultaneous reducing the temporal and spatial redundancy, ``Spatial$\&$Temporal" achieves the highest FR (100\%), and smallest QN and MAP.  {The average FR increases 26.7\% (73.3\%$\rightarrow$100\%), average QN and MAP decrease about 36\% (3662$\rightarrow$2227)  and 47\% (6.37$\rightarrow$3.35) versus the baseline, respectively.  In addition, ``Spatial$\&$Temporal" has smaller variance than baseline. For example, the variance has no changes for FR metric, and only changes 1\% around the mean for MAP metric,  4\% around the mean for NQ metric. For this reason,  we  neglect the variance value in the following comparison experiments. } This verifies the important role of dimension reduction when performing video attacks. 

\begin{figure}[t]
\center
\includegraphics[width=1\linewidth]{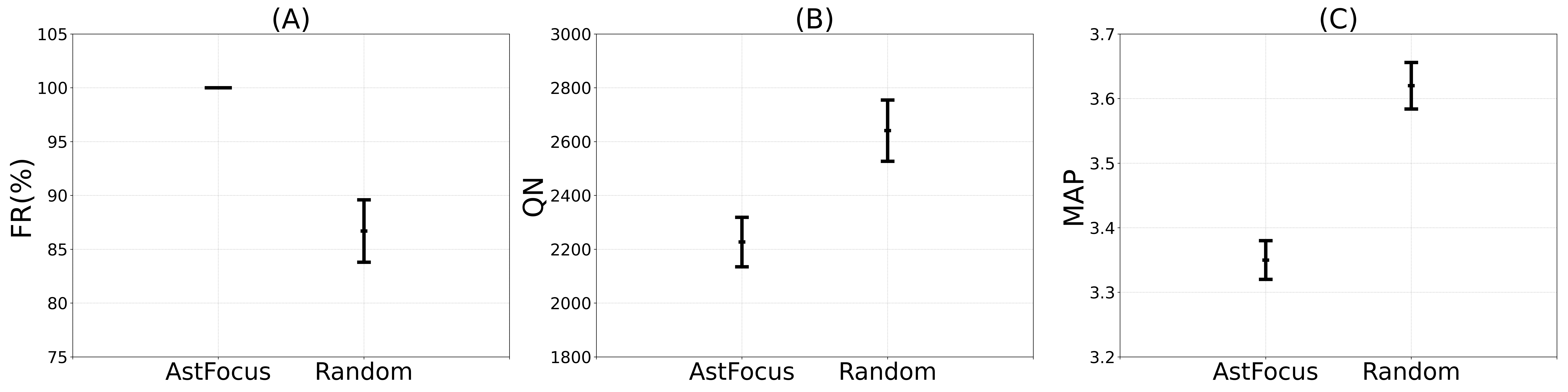}
\caption{ {Comparison between RL-based agent and  random agents. }}
\label{frandomrl12}
\end{figure}

We also compare our RL-based agents with random agents, where the key frames and key patches are randomly selected in each PGD iteration, other settings are the same. Figure \ref{frandomrl12} shows the comparison results.  {We can see that for all the evaluation metrics, our RL-based agent outperforms the random agent (100\%$\pm$0\% vs 86.7\%$\pm$2.89\%, 2227$\pm$92 vs 2632$\pm$114, 3.35$\pm$0.03 vs 3.62$\pm$0.04), which verifies the temporal and spatial agents in our method are jointly well-trained under the guidance of the carefully designed rewards.} With the feedback of threat models, the temporal and spatial agents become intelligent.

\subsubsection{Effects of various rewards}

To better guide the agent learning, we carefully design the rewards. Specifically, one common reward Eq.(\ref{eq:common}) coming from the black-box threat model, and two kinds of specific rewards Eq.(\ref{eq:edgebox}) as well as  Eq.(\ref{eq:rep}) and Eq.(\ref{eq:upper_bound}). The common reward is shared by the spatial agent and temporal agent, and the special rewards only belong to their own agents. In this section, we explore the effects of these rewards to the fooling rate, query number and perturbations.

\begin{table}[t]
\caption{ {Effects of various rewards to AstFocus attacks in an un-targeted setting.}} 
\vspace{-0.5cm}
\center 
\setlength{\tabcolsep}{1.3 mm}{
\begin{tabular}{ccccc}
\hline
\multirow{2}{*}{Metrics} & \multicolumn{4}{|c}{Different reward versions}                                                                   \\ \cline{2-5} 
                         & \multicolumn{1}{|c|}{Common}   & \multicolumn{1}{c|}{+Edgebox}    & \multicolumn{1}{c|}{+Sparse}   &+Representative   \\ \hline
FR(\%)                    & \multicolumn{1}{|c|}{ 76.7$\pm$2.9}   & \multicolumn{1}{c|}{91.7$\pm$2.9}    & \multicolumn{1}{c|}{96.7$\pm$2.9}   & 100$\pm$0.0    \\ \hline
QN                        & \multicolumn{1}{|c|}{3780$\pm$183}  & \multicolumn{1}{c|}{2690$\pm$122} & \multicolumn{1}{c|}{2490$\pm$87}  & 2227$\pm$92 \\ \hline
MAP                      & \multicolumn{1}{|c|}{3.62$\pm$0.07} & \multicolumn{1}{c|}{3.58$\pm$0.04}  & \multicolumn{1}{c|}{3.39$\pm$0.04} &3.35$\pm$0.03 \\ \hline
\end{tabular}}
\label{tab:ablation_results2}
\end{table}

Table \ref{tab:ablation_results2} lists the ablation study for effects of various rewards. The term ``Common" denotes the guidance of both temporal and spatial agents only using the common reward in Eq.(\ref{eq:common}). The terms ``+Edgebox", ``+Sparse", and `+Representative" denote adding the corresponding rewards (Eq.(\ref{eq:edgebox}), Eq.(\ref{eq:upper_bound}), and Eq.(\ref{eq:rep}), respectively) on the former basis to guide the agent learning. From the table, we see that with the addition of more and more rewards, average FR gradually increases, and average QN and average MAP gradually decrease.  {Compared with the solo common reward, the full version (the rightmost column) with all the rewards improves 23.3\% for average FR (76.7\%$\rightarrow$100\%), and reduces 43\% for average QN (3780$\rightarrow$2227), and 8\% for average  MAP (3.62$\rightarrow$3.35). The variance also becomes smaller and smaller.} The contrast verifies the rationality of the designed rewards.

\begin{table*}[t]
\center
\caption{ {The comparative results versus four different threat models on UCF-101 dataset.  The best results are highlighted in red. The symbol ``-" means the used NQ exceeds the maximum NQ.  $\uparrow$ denotes the larger, the better, and $\downarrow$ denotes the smaller, the better.}}
\label{tab:final2}
\setlength{\tabcolsep}{1.3 mm}{
\begin{tabular}{c|c|c|cccc|cccc}
\hline
\multirow{2}{*}{Datasets} & \multirow{2}{*}{Threat Models} & \multirow{2}{*}{Attack Methods} & \multicolumn{4}{c|}{Un-targeted attacks} & \multicolumn{4}{c}{Targeted attacks} \\ \cline{4-11} 
 &  &  & MAP $\downarrow$ & NQ $\downarrow$ & FR $\uparrow$ & T(s)$\downarrow$& MAP$\downarrow$ & NQ$\downarrow$ & FR $\uparrow$  &T(s)$\downarrow$\\ \hline
\multirow{26}{*}{UCF-101} & \multirow{6}{*}{TSM \cite{lin2019tsm}}

& VBAD attack \cite{jiang2019black} &6.023   &  {2803}   &84\%  &   {31.2}   &9.208   &15304   &63\%  & \textbf{201.7}  \\ 

                          &                      &Heuristic attack \cite{wei2019heuristic} &5.956 &10657   &40\% & {212.9}     &--  &--  &-- & {--}  \\ 
                          
                          &                      &Sparse attack \cite{wei2022sparse} &\textbf{3.417}   &8529   &58\% & {147.2}    &--  &--  &-- & {--}  \\ 
                          
                          &                      &Motion-sampler attack \cite{zhang2020motion} &7.237   &5187  &83\% & {124.6}    &7.415   &13577  &79\% & {288.6}   \\ 
                          &                      & GEO-TRAP attack \cite{li2021adversarial} &5.865   &3782  &   {88\%}  & {87.2}   &  {6.265}   &  {11494}  &  {84\%}  &   {247.6}  \\ 
                           &                      & RLSB attack \cite{wang2021reinforcement}  &4.823        &4898        & {87\%}   & {101.3}      &7.274      &20532      &40\%  & {365.4}  \\ 
                          &                      &AstFocus attack (ours) &   {3.355}   &\textbf{1138}   &\textbf{96\%} & \textbf{24.6}    &\textbf{4.546}  &\textbf{8064}   &\textbf{100\%}& {274.5}   \\ \cline{2-11}
                          
                          & \multirow{6}{*}{TSN \cite{wang2016temporal}} 
                          
                          & VBAD attack \cite{jiang2019black} &6.168  &  {2450}   &84\% & \textbf{13.8}    &9.391  &18960   &47\% & {394.6}  \\ 
                          
                          &                       & Heuristic attack \cite{wei2019heuristic} &5.265  &9135   &51\% & {141.7}    &--   &--   &-- & {--}   \\ 
                          
                          &                       & Sparse attack \cite{wei2022sparse} &\textbf{3.131}   &6916   &64\% & {181.4}    &--   &--   &-- & {--}  \\ 
                          
                          &                       & Motion-sampler attack \cite{zhang2020motion} &6.895   &4744   &78\%  & {95.6}   &6.903  &19626  &  {59\%}  & {316.8}  \\ 
                          
                          &                       & GEO-TRAP attack \cite{li2021adversarial} &5.472   &3782  & {75\%}  & {87.3}   &  {5.952}   &  {18585}   &55\%  &   {301.2}  \\ 
                           &                      & RLSB attack \cite{wang2021reinforcement}  &5.238      &3504       &   {93\%}   & {48.2}      &8.448      &  20668     &44\%  & {289.5}   \\ 
                          &                       & AstFocus attack (ours) &   {3.265}   &\textbf{2015}   &\textbf{99\%}  &   {37.4}   &\textbf{4.495}   &\textbf{8483}   &\textbf{76\%}  & \textbf{272.9}  \\ \cline{2-11} 
                          
& \multirow{6}{*}{C3D \cite{hara2018can}}  

& VBAD attack \cite{jiang2019black} & 6.800    &  {4890}      & {75\%}  & \textbf{43.4}      &10.760  &20234  &60\% & \textbf{139.2}      \\ 

                          &                      & Heuristic attack \cite{wei2019heuristic} &6.295  &14160  &30\% & {143.2}     &--  &--  &--   & {--}             \\ 
                          
                          &                      & Sparse attack \cite{wei2022sparse} &\textbf{3.009}       &9507       &42\%    & {86.0}       &--       &--       &--    & {-- }       \\ 
                          &                      & Motion-sampler attack \cite{zhang2020motion} &6.153      &8132        &62\%  & {97.9}      &7.242         &20690        &47\%  &252.9     \\ 
                          &                      & GEO-TRAP attack \cite{li2021adversarial} &5.877        &7045         &  {75\%} & {74.4}        & {6.332}      &  {17340}      & {77\%} &  {205.4}   \\ 
                                         &                      & RLSB attack \cite{wang2021reinforcement}  &5.326        &6568      & {68\%}   & {72.0}      &7.225     &22018       &35\%  &207.1    \\ 
                                         
                          &                      & AstFocus attack (ours) &  {4.015}   &\textbf{4224}   &\textbf{90\%}  &  {66.8}   &\textbf{4.225}   &\textbf{13470}   &\textbf{88\%}   & 236.4  \\ \cline{2-11} 
                          
                          & \multirow{6}{*}{SlowFast \cite{feichtenhofer2019slowfast}} 
                          
                          & VBAD attack \cite{jiang2019black} &6.302   &  {4089}   &77\% &  {42.5}    &9.118  &19423   &53\% & {499.2}  \\ 
                          
                          &                       & Heuristic attack \cite{wei2019heuristic} &5.869   &12776   &34\% & {260.5}    &--  &--  &-- & {--}  \\ 
                          
                         &                       & Sparse attack \cite{wei2022sparse} &\textbf{3.164}   &8642   &58\% & {168.8}    &--  &--  &-- & {--}  \\ 
                         
                         &                       & Motion-sampler attack \cite{zhang2020motion} &7.086  &5166   & {77\%}  & {136.3}   &7.275   & 17928 &56\% & {451.8}   \\ 
                         &                       & GEO-TRAP attack \cite{li2021adversarial} &5.712   &4334   &84\% & {120.5}    & {6.273}   & {16506}   & {62\%} & {532.1}   \\ 
                          &                      & RLSB attack \cite{wang2021reinforcement}  &5.586        &4563        &  {85\%} & {94.5}        &7.655       & 22552      &43\%  &  {448.4}  \\ 
                         &                       & AstFocus attack (ours) &  {4.286}   &\textbf{1435}   &\textbf{93\%} & \textbf{35.7}    &\textbf{4.436}   &\textbf{13660}   &\textbf{85\%} & \textbf{385.3}   \\ \hline
                         
\end{tabular}
}\vspace{-0.4cm}
\end{table*}

\begin{table*}[t]
\center
\caption{ {The comparative results versus four different threat models on HMDB-51 dataset.  The best results are highlighted in red. The symbol ``-" means the used NQ exceeds the maximum NQ.  $\uparrow$ denotes the larger, the better, and $\downarrow$ denotes the smaller, the better.}}
\label{tab:final3}
\setlength{\tabcolsep}{1.3 mm}{
\begin{tabular}{c|c|c|cccc|cccc}
\hline
\multirow{2}{*}{Datasets} & \multirow{2}{*}{Threat Models} & \multirow{2}{*}{Attack Methods} & \multicolumn{4}{c|}{Un-targeted attacks} & \multicolumn{4}{c}{Targeted attacks} \\ \cline{4-11} 
 &  &  & MAP $\downarrow$ & NQ $\downarrow$ & FR $\uparrow$ & T(s)$\downarrow$& MAP$\downarrow$ & NQ$\downarrow$ & FR $\uparrow$  &T(s)$\downarrow$\\ \hline
\multirow{26}{*}{HMDB-51} & \multirow{6}{*}{TSM \cite{lin2019tsm}}

& VBAD attack \cite{jiang2019black} &6.361   & {1818}   &92\%  & \textbf{22.4}   &9.057   &17550  &60\%  & {495.6}  \\ 

                          &                      &Heuristic attack \cite{wei2019heuristic} &5.043   &10385   &58\% & {211.4}     &--  &--  &-- & {--}  \\ 
                          
                          &                      &Sparse attack \cite{wei2022sparse} &\textbf{3.334}   &6244   &62\% & {102.5}    &--  &--  &-- & {--}  \\ 
                          
                          &                      &Motion-sampler attack \cite{zhang2020motion} &7.229   &3911   &90\% & {95.8}    &8.012   &19508  &58\% & {686.7}   \\ 
                          
                          &                      &GEO-TRAP attack \cite{li2021adversarial} &5.919   &3164   & {92\%}  & {84.8}   &{6.222}   &10836   &84\% & \textbf{337.7}  \\ 
                          
                           &                      & RLSB attack \cite{wang2021reinforcement}  &5.323        &5950         & {82\%}   & {112.9}      & 7.754      &20171       &28\%  & {575.8}  \\ 
                           
                          &                      &AstFocus attack (ours) & {3.411}   &\textbf{1529}   &\textbf{100\%} & {34.7}    &\textbf{4.326}   &\textbf{7319}   &\textbf{92\%}& {419.1}   \\ \cline{2-11}
                          
                          & \multirow{6}{*}{TSN \cite{wang2016temporal}} & VBAD attack \cite{jiang2019black} &5.873   & {2373}   &90\% & \textbf{26.4}    &9.244  &21795  &46\% & {659.1}  \\ 
                          
                          &                       & Heuristic attack \cite{wei2019heuristic} &5.395   &10146   &58\% & {172.5}    &--   &--   &-- & {--}   \\ 
                          
                          &                       & Sparse attack \cite{wei2022sparse} &\textbf{3.271}   &6765   &74\% & {105.9}    &--   &--   &-- & {--}  \\ 
                          
                          &                       & Motion-sampler attack \cite{zhang2020motion} &7.275   &3667   &88\%  & {74.1}   &8.087  &24332  &28\%  & {964.6}  \\ 
                          
                          &                       & GEO-TRAP attack \cite{li2021adversarial} &5.192   &3392   & {88\%}  & {61.6}   & 6.344  &21492   &36\%  & {744.6}  \\ 
                          
                           &                      & RLSB attack \cite{wang2021reinforcement}  &5.312        &4217         & {92\%}   & {68.6}      & 6.494      & 22718      &22\%  & {719.1}   \\ 
                           
                          &                       & AstFocus attack (ours) & {3.52}   &\textbf{2198}   &\textbf{96\%}  & {51.1}   &\textbf{4.090}   &\textbf{9953}   &\textbf{74\%}  &\textbf{522.4}  \\ \cline{2-11} 
                          
& \multirow{6}{*}{C3D \cite{hara2018can}} 
& VBAD attack \cite{jiang2019black} &6.743   & {4107}   &78\%  &\textbf{36.5}   &10.528   &22302   &64\%  & \textbf{361.2}   \\ 

                          &                      &Heuristic attack \cite{wei2019heuristic} &4.838   &10534   &42\% & {117.4}     &--  &--  &-- & {--}  \\ 
                          
                          &                      &Sparse attack \cite{wei2022sparse} &\textbf{2.983}   &8545   &46\% & {73.8}    &--  &--  &-- & {--}  \\ 
                          
                          &                      &Motion-sampler attack \cite{zhang2020motion} &7.035   &6491   &68\% & {89.5}    &7.973   &22199   &44\% & 558.4   \\ 
                          
                          &                      &GEO-TRAP attack \cite{li2021adversarial} &5.666   &5082   & {84\%}  & {48.2}   &{6.324}   &16374   &74\% &518.2   \\ 
                          
                           &                      & RLSB attack \cite{wang2021reinforcement}  &4.688        &7279         & {62\%}   & {77.1}      &7.212      &18190     &60\%  &530.8   \\ 
                           
                          &                      &AstFocus attack (ours) & {3.835}   &\textbf{3628}   &\textbf{92\%} & {45.6}    &\textbf{4.025}   &\textbf{9997}   &\textbf{86\%}& {473.6}   \\ \cline{2-11}
                          
                          & \multirow{6}{*}{SlowFast \cite{feichtenhofer2019slowfast}} 
                          & VBAD attack \cite{jiang2019black} &6.528   & {5442}   &72\% & \textbf{46.7}    &10.615   & 22955  &36\% & {693.3}  \\ 
                          
                          &                       & Heuristic attack \cite{wei2019heuristic} &5.875   &9094   &54\% & {162.9}    &--  &--  &-- & {--}  \\ 
                          
                         &                       & Sparse attack \cite{wei2022sparse} &\textbf{3.228}   &8977   &56\% & {138.0}    &--  &--  &-- & {--}  \\ 
                         
                         &                       & Motion-sampler attack \cite{zhang2020motion} &7.163   &6553   & {74\%}  & {156.4}   &7.956   &18513   &52\% & {652.3}   \\ 
                         
                         &                       & GEO-TRAP attack \cite{li2021adversarial} &6.242   &5741   &78\% & {122.8}    &{6.179}   &{17636}   &{46\%} & {666.1}   \\ 
                         
                          &                      & RLSB attack \cite{wang2021reinforcement}  &5.68        &4495         & {84\%} & {81.9}        &7.416       &20378      &34\%  & {650.0}  \\ 
                          
                         &                       & AstFocus attack (ours) & {4.078}   &\textbf{2295}   &\textbf{96\%} & {47.1}    &\textbf{4.682}   &\textbf{13970}   &\textbf{78\%} & \textbf{567.9}   \\ \hline
\end{tabular}
}\vspace{-0.3cm}
\end{table*}

\subsubsection{Convergence of AstFocus attacks}

\begin{figure}[t]
\center
\includegraphics[width=0.8\linewidth]{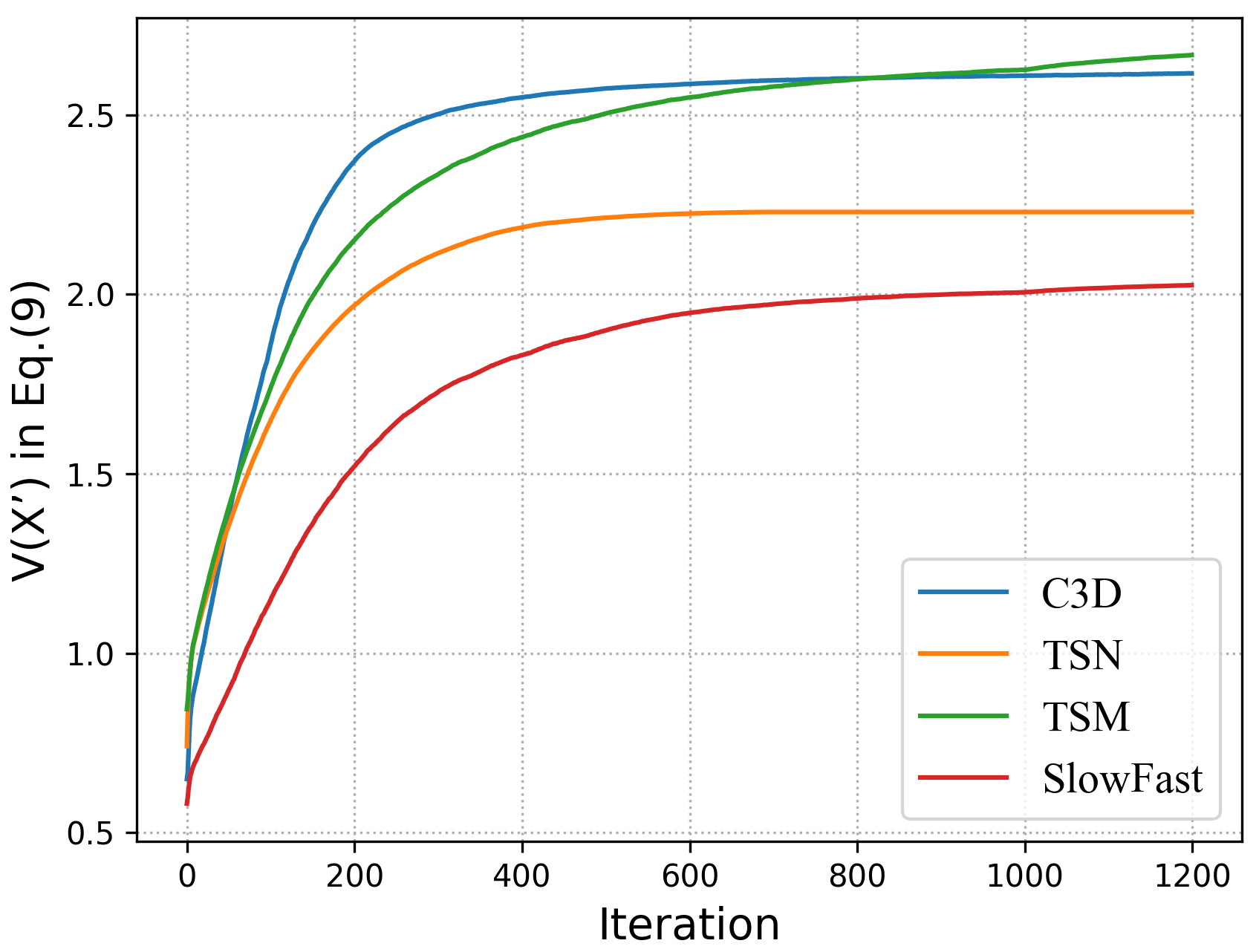}
\vspace{-0.3cm}
\caption{ {The convergence of the proposed AstFocus attacks.} }
\label{fig:converge}
\end{figure}

\begin{figure}[t]
\center
\includegraphics[width=1\linewidth]{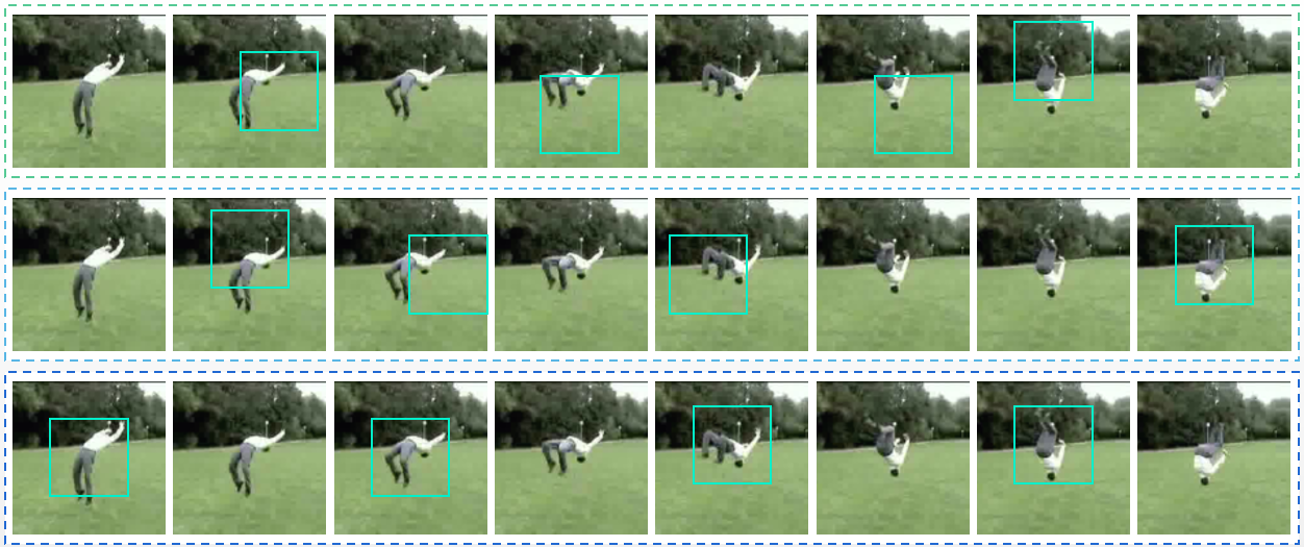}
\caption{A qualitative example for selecting key frames and key regions in  AstFocus attacks.  From top to bottom denotes the selected key frames and  key patches in the 5-th, 10-th, and 20-th PGD iteration. }
\label{fig:converge_quali}
\end{figure}

Because our agents are updated by the rewards in each iteration, it is necessary to investigate whether the agents are under the convergence with the increasing iterations.  {For that, we list the values' change of Eq.(\ref{eq:common1}) with the increasing PGD iteration in Figure \ref{fig:converge}. Eq.(\ref{eq:common1}) directly reflects the success or failure of an attack. If the target class's confidence score is above the ground-truth class's confidence score, the value of Eq.(\ref{eq:common1}) will be above 1, representing that the attack is successful, and vice versa. From the figure, we can see that  Eq.(\ref{eq:common1})'s values for all the threat models are gradually increasing until the stable situation. When the iteration reaches 400, all the models achieves the convergence. This verifies the good convergence of  AstFocus attack. Actually, the attack usually stops when Eq.(\ref{eq:common1})'s value is above 1, i.e., the step 9 in Algorithm \ref{alg::conjugateGradient}. Therefore, we only need few iterations in application.} Figure \ref{fig:converge_quali} gives a qualitative example of the agents in AstFocus attacks, where the key frames and key patches in different iterations are illustrated by the bounding boxes. We can see that the spatial agent gradually focuses on the foreground objects. This is reasonable because these areas are key cues for video recognition task. In addition, the temporal agent tends to select the frames with big changes in the actions. These frames have a strong representative ability for the whole video from the appearance, which shows key frames are sensitive to attacks.

\begin{table*}[t]
\center
\caption{ {The comparative results versus four different threat models on Kinetics-400 dataset.  The best results are highlighted in red. The symbol ``-" means the used NQ exceeds the maximum NQ.  $\uparrow$ denotes the larger, the better, and $\downarrow$ denotes the smaller, the better.}}
\label{tab:final4}
\setlength{\tabcolsep}{1.3 mm}{
\begin{tabular}{c|c|c|cccc|cccc}
\hline
\multirow{2}{*}{Datasets} & \multirow{2}{*}{Threat Models} & \multirow{2}{*}{Attack Methods} & \multicolumn{4}{c|}{Un-targeted attacks} & \multicolumn{4}{c}{Targeted attacks} \\ \cline{4-11} 
 &  &  & MAP $\downarrow$ & NQ $\downarrow$ & FR $\uparrow$ & T(s)$\downarrow$& MAP$\downarrow$ & NQ$\downarrow$ & FR $\uparrow$  &T(s)$\downarrow$\\ \hline
\multirow{26}{*}{Kinetics-400} & \multirow{6}{*}{TSM \cite{lin2019tsm}}
& VBAD attack \cite{jiang2019black} & 6.480    & {3626}      & {78\%}  & \textbf{16.8}      &10.338 &23670 &34\% & {593.6}     \\ 

                          &                      & Heuristic attack \cite{wei2019heuristic} &5.744  &11918  &56\% &202.1      &--  &--  &--   & {--}             \\ 
                          
                          &                      & Sparse attack \cite{wei2022sparse} &\textbf{2.725}       &9105       &60\%    &147.6        &--       &--       &--    & {-- }       \\ 
                          
                          &                      & Motion-sampler attack \cite{zhang2020motion} &7.234      &4494        &88\%  & {105.9}      &8.033        &21032     &46\%  & {803.1}    \\ 
                          
                          &                      & GEO-TRAP attack \cite{li2021adversarial} &6.007        &3962         & {92\%} & {87.1}        &6.217     &15585      &72\% & \textbf{536.0}   \\ 
                                         &                      & RLSB attack \cite{wang2021reinforcement}  &5.856        &4422         & {84\%}   & {91.4}      &7.200       &21724       &26\%  & {653.2}   \\ 
                                         
                          &                      & AstFocus attack (ours) & {3.658}   &\textbf{2416}   &\textbf{96\%}  & {44.1}   &\textbf{4.482}   &\textbf{9758}   &\textbf{88\%}   & {556.5}  \\ \cline{2-11} 
                          
                          & \multirow{6}{*}{TSN \cite{wang2016temporal}} 
                          & VBAD attack \cite{jiang2019black} &5.764   & {1668}   &92\% & \textbf{22.6}    &9.414   &17560   &58\% &\textbf{427.9}   \\ 
                          
                          &                       & Heuristic attack \cite{wei2019heuristic} &4.806   &10080   &52\% & {165.6}    &--   &--   &-- & {--}   \\ 
                          
                          &                       & Sparse attack \cite{wei2022sparse} &\textbf{2.579}   &8212  &54\% &117.8     &--   &--   &-- & {--}  \\ 
                          
                          &                       & Motion-sampler attack \cite{zhang2020motion} &6.982   &3422   &90\%  & {80.3}   &7.979  &22119  &36\%  & {735.2}  \\ 
                          
                          &                       & GEO-TRAP attack \cite{li2021adversarial} &5.636   &2684   & {90\%}  & {53.1}   &5.789   &16402   &50\%  & {459.7}  \\ 
                          
                           &                      & RLSB attack \cite{wang2021reinforcement}  &5.395        &3774         & {94\%}   & {63.7}      &7.140       & 23574      &22\%  & {622.4}   \\ 
                           
                          &                       & AstFocus attack (ours) & {3.349}   &\textbf{1021}   &\textbf{100\%}  & {26.7}   &\textbf{4.684}   &\textbf{9940}   &\textbf{90\%}  & {558.4}  \\ \cline{2-11} 
                          
& \multirow{6}{*}{C3D \cite{hara2018can}} 
& VBAD attack \cite{jiang2019black} &5.640   & {3444}   &90\%  & {13.2}   &10.230   &22070   &52\%  & {135.2}  \\ 

                          &                      &Heuristic attack \cite{wei2019heuristic} &5.805   &  11384&48\% & {44.7}     &--  &-- &-- & {--}  \\ 
                          
                          &                      &Sparse attack \cite{wei2022sparse} &\textbf{2.769}  &5045   &78\% & 20.7   &--  &--  &-- & {--}  \\ 
                          
                          &                      &Motion-sampler attack \cite{zhang2020motion} &6.895   &2485   &96\% & {14.1}    &7.808   &15679   &70\% & {163.2}   \\ 
                          
                          &                      &GEO-TRAP attack \cite{li2021adversarial} &6.135   & 3436  &96{\%}  & {15.6}   &6.334{}   &12260   &90\% & \textbf{120.5}  \\ 
                          
                           &                      & RLSB attack \cite{wang2021reinforcement}  & 4.925       &5915         & 76{\%}   & {25.2}      &8.053       &20975       &36\%  & {179.6}  \\ 
                           
                          &                      &AstFocus attack (ours) & {3.858}   &\textbf{1055}  &\textbf{100\%} & \textbf{9.1}    &\textbf{4.728}   &\textbf{10840}   &\textbf{92\%}& {152.9}   \\ \cline{2-11}
                          
                          & \multirow{6}{*}{SlowFast \cite{feichtenhofer2019slowfast}} & VBAD attack \cite{jiang2019black} &6.667   & {2732}   &86\% & {33.6}    &10.532   &19970  &44\% & {599.2}  \\ 
                          
                          &                       & Heuristic attack \cite{wei2019heuristic} &4.723   &9154   &54\% & {159.5}    &--  &--  &-- & {--}  \\ 
                          
                         &                       & Sparse attack \cite{wei2022sparse} &\textbf{3.043}   &5901   &70\% &97.8     &--  &--  &-- & {--}  \\ 
                         
                         &                       & Motion-sampler attack \cite{zhang2020motion} &7.145   &2282   & {92\%}  & {55.4}   &7.953   &20264   &40\% & {878.3}   \\ 
                         
                         &                       & GEO-TRAP attack \cite{li2021adversarial} &5.832   &1646   &94\% & {37.5}    &{6.197}   &{9594}   &{86\%} & {366.8}   \\ 
                         
                          &                      & RLSB attack \cite{wang2021reinforcement}  &4.636        &4802         & {88\%} & {88.1}        & 7.405      & 21137      &34\%  & {705.4}  \\ 
                          
                         &                       & AstFocus attack (ours) & {3.356}   &\textbf{851}   &\textbf{100\%} & \textbf{24.4}    &\textbf{4.015}   &\textbf{7572}   &\textbf{98\%} & \textbf{386.2}   \\ \hline
\end{tabular}
}\vspace{-0.3cm}
\end{table*}

\subsection{Comparisons with SOTA methods}
Here, we compare the proposed AstFocus attack with six state-of-the-art black-box video attack methods on three public datasets and four widely used video recognition models. The comparative results in the un-targeted and targeted settings are recorded in Table \ref{tab:final2}, Table \ref{tab:final3}, and \ref{tab:final4} (for fair comparison, the target label for all the methods are the same when performing targeted attacks).  {From the tables, we  see that: \textbf{(1)} \textbf{For attack effect (FR and MAP)}, our method significantly outperforms other six SOTA methods for FR metric (at least 5\%) versus all the threat models on all the datasets, showing the big advantage in attacking ability. For MAP metric, AstFocus attack only slightly loses to Sparse attack in the un-targeted attack but obviously outperforms other five video attacks. Because Sparse attack adds adversarial perturbations only on the fixed key frames in each PGD iteration. In Eq.(\ref{eq:pgd}), there exists a clip operation $Proj(\cdot)$ to project the perturbations to a small range. So the upper bound of adversarial perturbations generated by Sparse attack is small. This design also limits the attacking efficiency and effectiveness, for example, Sparse attack only has almost 40\% FR but needs almost 9000 NQ on average for un-targeted attacks, far less than AstFocus attacks. Overall, AstFocus attack is better than Sparse attack. A small MAP under a high FR means an accurate evaluation for the models' adversarial robustness.  From this viewpoint, AstFocus attack is more suitable to  evaluate different video models.  \textbf{(2)} \textbf{For attack efficiency (NQ and T)}, AstFocus also significantly beats other six SOTA methods for NQ versus all the threat models on all the datasets, reducing at least 10\% queries compared with the second best video attacks. For time metric, AstFocus attack only slightly loses to VBAD attack but still beats other five video attacks. This is reasonable because AstFocus attack integrates two additional  agents to reduce dimensions during attacks while VBAD does not involve this step. In return, AstFocus greatly outperforms VBAD versus the other three metrics. Overall, AstFocus has the high efficiency.  \textbf{(3)} \textbf{For simultaneous modeling}, AstFocus attack remarkably outperforms RLSB attack on all the settings, showing simultaneously modeling the key frames and key regions is indeed more effective than separately modeling them. This also demonstrates the core idea in this paper. } \textbf{(4)} \textbf{From the view of robustness evaluation}, all the seven black-box video attacks show C3D has better adversarial robustness than the other models. The C3D has lower FR values but higher NQ values, which shows C3D is harder to attack. This may motivates us an in-depth study for the C3D's structure to design robust video recognition models. 


\begin{figure}[t]
\center
\includegraphics[width=1\linewidth]{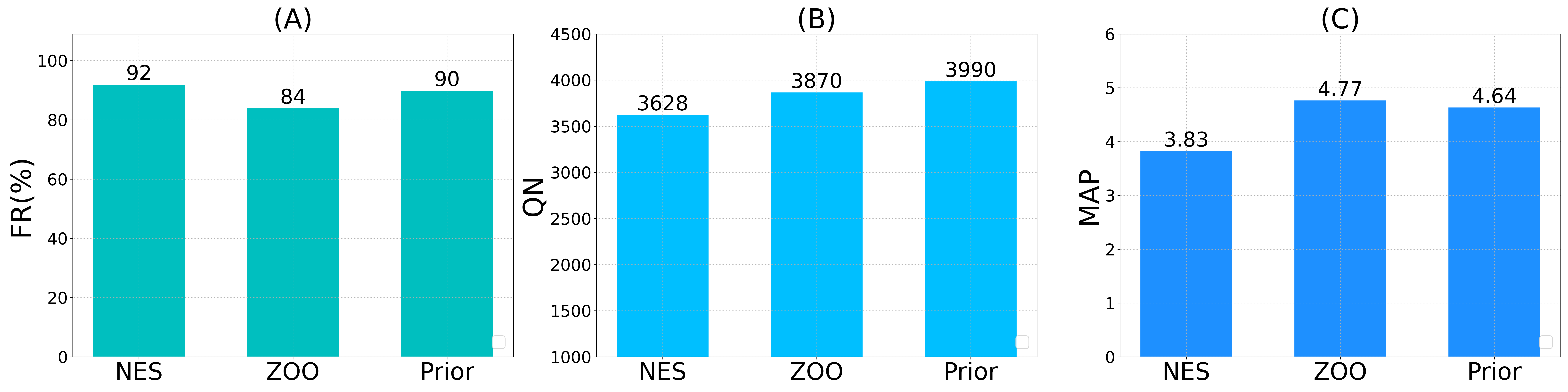}
\caption{Comparisons of different gradient estimators within AstFocus. }
\label{fig:othergrad}
\end{figure}

\begin{figure*}[t]
\center
\includegraphics[width=0.48\linewidth]{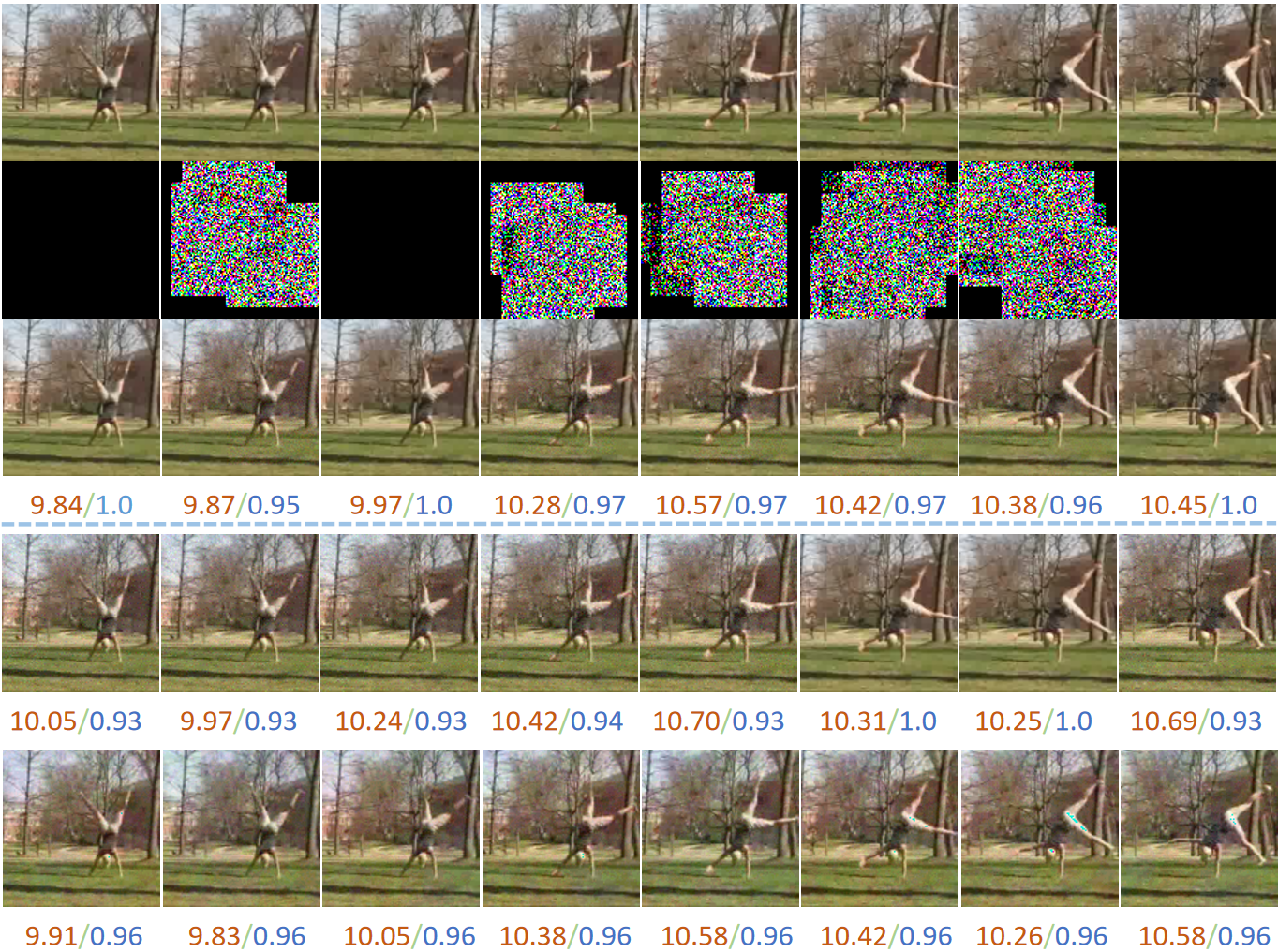}
\includegraphics[width=0.483\linewidth]{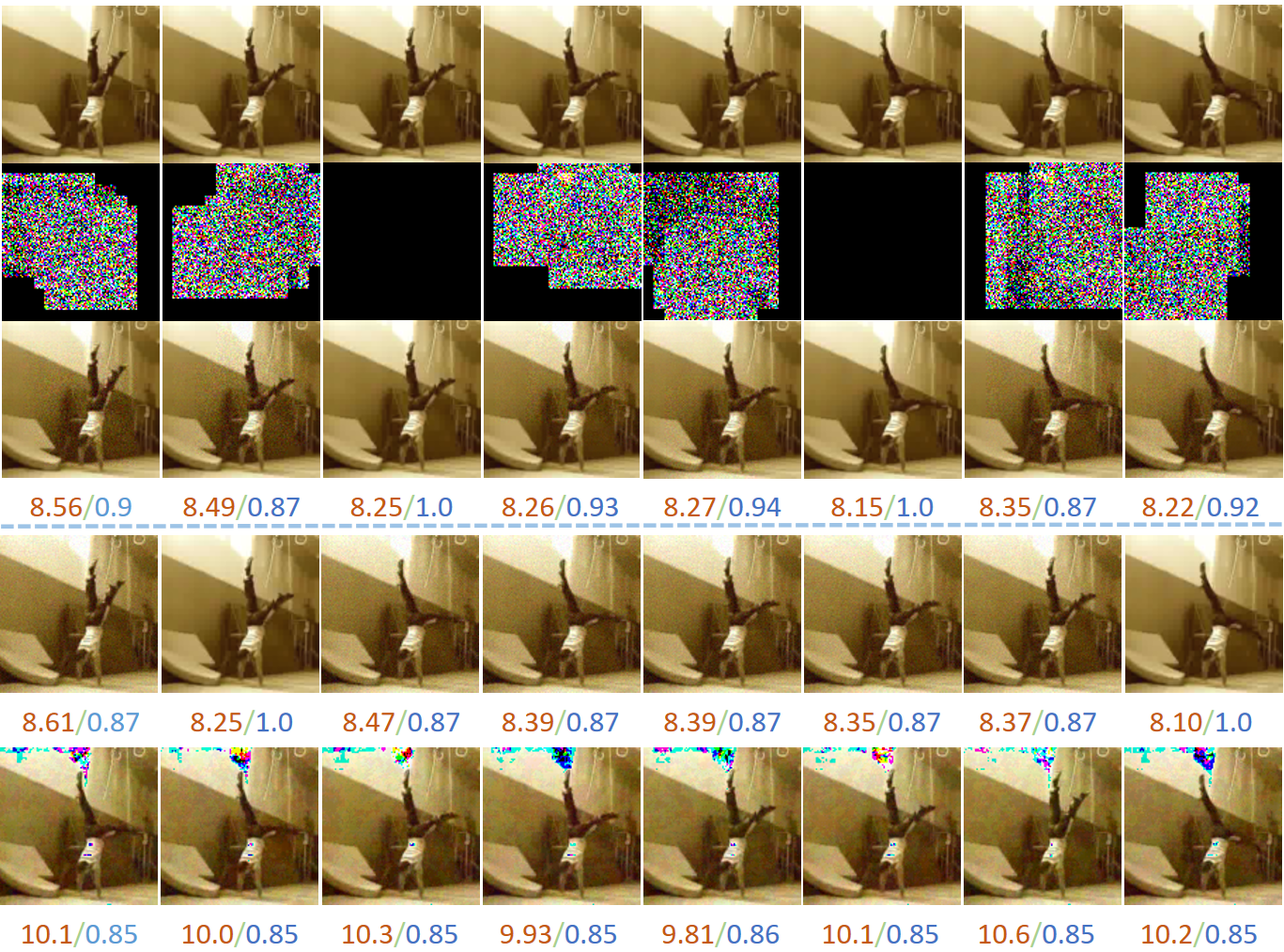}
\caption{ {Two qualitative examples output by AstFocus attacks. For each example, from top to bottom rows are clean video, adversarial perturbations, and our adversarial video, respectively. Two adversarial videos generated by RLSB attack and GEO-TRAP attack are listed below the dotted line as a reference. We compute two metrics as (blurriness/SSIM) for each video frame. For the left blurriness degree \cite{marziliano2002no}, the smaller the better, and for the right SSIM \cite{hore2010image}, the larger the better.}}
\label{fig:final_quali}
\end{figure*}

\begin{table}[h]
\caption{ {Results of AstFocus attack against defended C3D method on HMDB-51}}
\centering 
\scalebox{0.8}{
\begin{tabular}{c|c|c|c|c}
\hline
\multicolumn{1}{l|}{Metrics} & \multicolumn{1}{c|}{No defense} & \multicolumn{1}{c|}{PGD-AT \cite{madry2017towards}}                      & \multicolumn{1}{c|}{OUD \cite{lo2021overcomplete}}                  & \multicolumn{1}{c}{AdvIT \cite{xiao2019advit}} \\ \hline
FR(\%)             &92.0          & 60.0 ($\downarrow$32)     & 72.0 ($\downarrow$20)     &70.0 ($\downarrow$22)                                   \\ \hline
QN                 &3628          & 7005 ($\uparrow$3377)  & 6283 ($\uparrow$2655)    & 2802 ($\downarrow$826)                                  \\ \hline
MAP                &3.835          & 4.814 ($\uparrow$0.979) & 5.090 ($\uparrow$1.255)  &  3.512 ($\downarrow$0.323)                                  \\ \hline
\end{tabular}}
\label{tab:defense}
\vspace{-0.3cm}
\end{table}

\subsection{Integrated with other gradient estimators}
In our AstFocus attack, the current gradient estimator is NES. Actually, we can replace NES with other state-of-the-art gradient estimators. To test this point, we conduct experiments. Here we choose two SOTA gradient estimators: Prior convictions  \cite{ilyas2018prior} and ZOO \cite{chen2017zoo}. Figure \ref{fig:othergrad} gives the results. We can see that when the gradient estimators are changed, the fooling rate, query number, and perturbation magnitudes only show a slight variation. Relatively speaking, NES achieves the better performance versus three metrics.  AstFocus attack is a flexible framework, which implies other modules can be replaced except the MARL module. The PGD can also be replaced with its improved versions.

\subsection{Qualitative results of AstFocus attacks}
We list two adversarial videos and the perturbations generated by AstFocus attacks in Figure \ref{fig:final_quali}, we see adversarial video is consistent with original video from the appearance, showing the imperceptibility of adversarial perturbations. To understand the perturbations, we enlarge their values to give a display. We see the final adversarial perturbations are sparse both in inter-frames and intra-frames. They show a superposition phenomenon by many noise patches  generated in each PGD iteration. These adversarial perturbations cover the foreground regions in key frames.  {We also give two adversarial videos generated by other recently published attack methods (RLSB attack and  GEO-TRAP attack) as a reference, where we can see our method has better imperceptible perturbations than the other methods.}

To better show the advantage, we compute two metrics to quantitatively measure the image quality. The first metric is to measure the blurriness degree \cite{marziliano2002no}. For this metric, the smaller the better. And another metric is SSIM \cite{hore2010image}. For this metric, the larger the better. We list these two metric values below each video frame as (blurriness/SSIM), where we see that our adversarial videos show better image quality than RLSB and GEO-TRAP attacks. 

\subsection{AstFocus attack against defense methods}
 {We evaluate the  performance of AstFocus attack against defense methods. Three kinds of representative video defense methods are chosen: Adversarial Training method (PGD-AT \cite{madry2017towards}), modifying network architecture method (OUD \cite{lo2021overcomplete}), and pre-processing method (AdvIT\footnote{AdvIT is proposed to detect the adversarial example. To adopt it to perform defense, we attach it before the threat model. If the input is detected as adversarial example, it will not be fed into the threat model. For this reason, the QN and MAP may decrease rather than increase.} \cite{xiao2019advit}).  The results for C3D model on HMDB-51 dataset are reported in Table \ref{tab:defense}, where the changes compared with the un-defended C3D are listed in the brackets. We can see that both the attacking performance and efficiency decrease. Specifically, the maximum drop of FR after defense is 32\%,  QN increases by 3377 at most, and MAP increases by 33\% at most. This is reasonable because the defended model will be harder to attack, but the FR, QN, and MAP are still acceptable. This  shows that AstFocus attack is  effective to evaluate the adversarial robustness even for the defended action recognition models. }

\section{Conclusion}
In this paper,  we designed the novel adversarial spatial-{t}emporal {focus}  attack on videos to  simultaneously identify the key frames and key regions in the video. AstFocus attack was based on the cooperative multi-agent reinforcement learning  framework. One agent was responsible for selecting key frames, and another agent was responsible for selecting key regions. These two agents were jointly trained by the common rewards received from the black-box threat models. By continuously querying, the reduced space composed of key frames and key regions was becoming precise, and the whole query number was less than that on the original video.  Extensive experiments on four famous video recognition models and three public action recognition datasets verified our efficiency and effectiveness, which was prevenient in fooling rate, query number, time, and perturbation magnitude at the same.

\section*{Acknowledgment}
This work is supported by National Key R$\&$D Program of China (Grant No.2020AAA0104002), National Natural Science Foundation of China (No.62076018).

\ifCLASSOPTIONcaptionsoff
  \newpage
\fi

\bibliographystyle{IEEEtran}
\bibliography{egbib}

\begin{IEEEbiography}[{\includegraphics[width=1in,height=1.25in,clip,keepaspectratio]{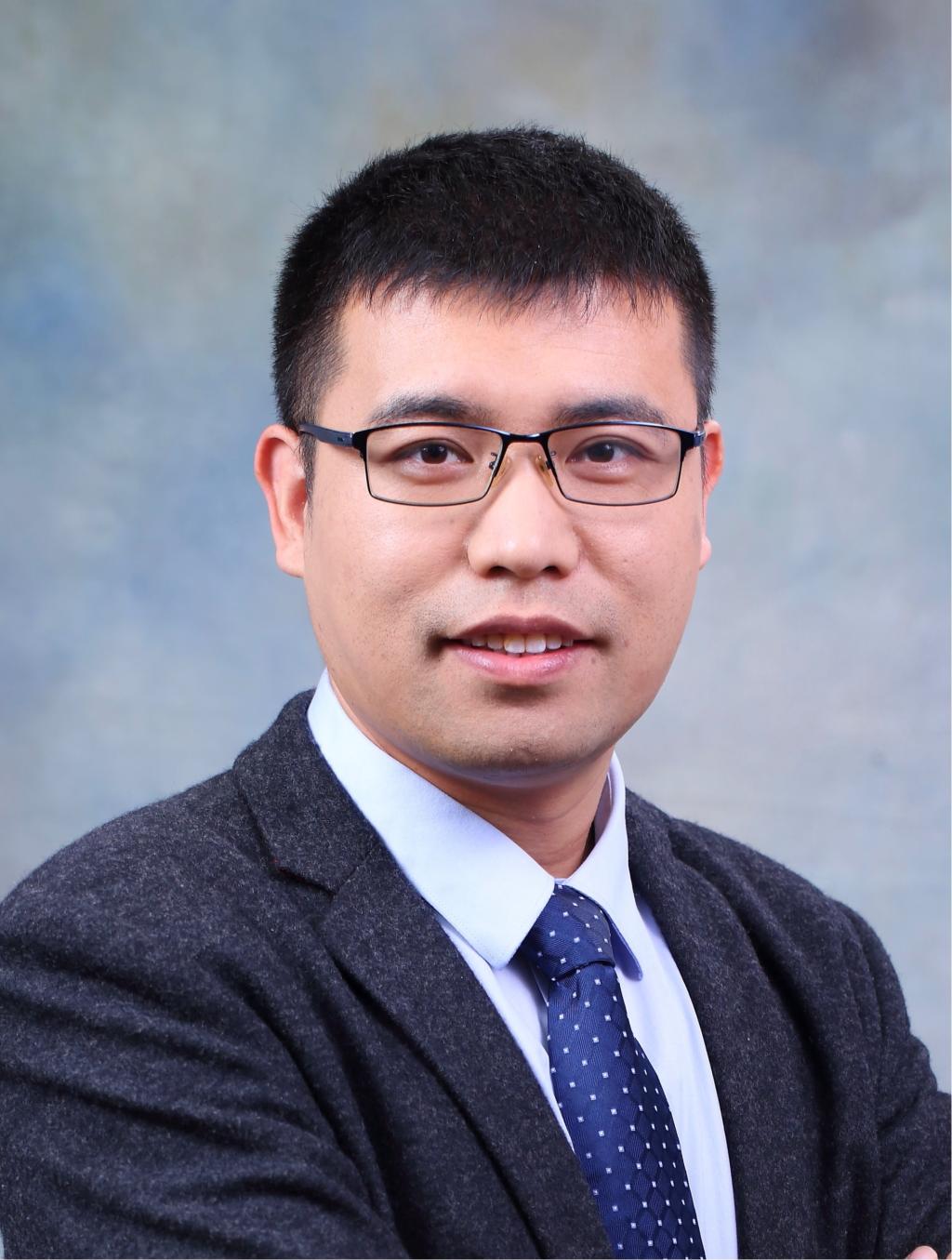}}]{Xingxing Wei}
received his Ph.D degree in computer science from Tianjin University, and B.S. degree in Automation from Beihang University, China. He is now an Associate Professor in Beihang University (BUAA). His research interests include computer vision, adversarial machine learning and its applications to multimedia content analysis. He is the author of referred journals and conferences in IEEE TPAMI, TMM, TCYB, TGRS, IJCV, PR, CVIU,  CVPR, ICCV, ECCV, ACMMM, AAAI, IJCAI etc.
\end{IEEEbiography}
\begin{IEEEbiography}[{\includegraphics[width=1in,height=1.25in,clip,keepaspectratio]{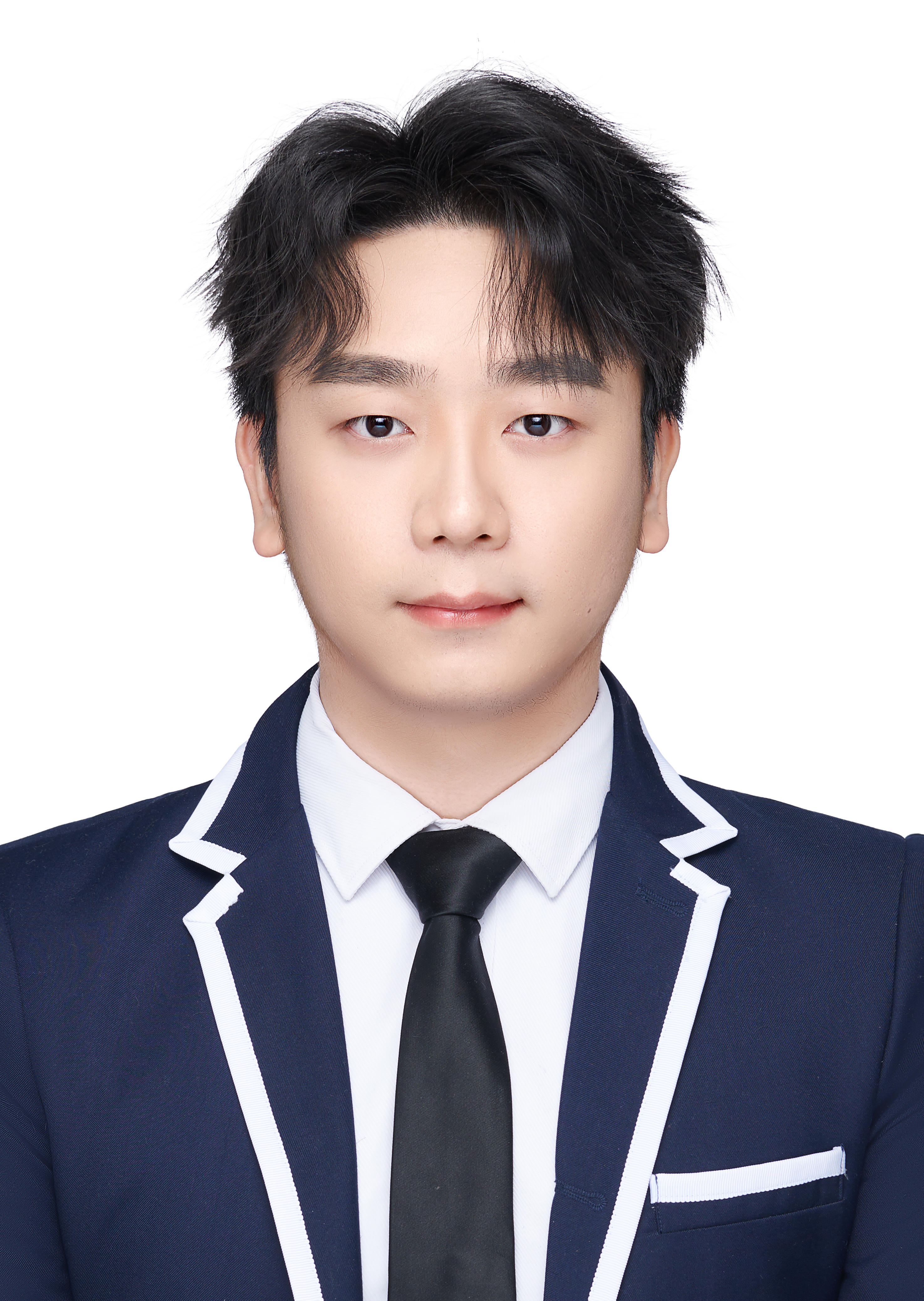}}]{Songping Wang} is now a Master student at school of software, Beihang University (BUAA).
His research interests include deep learning and adversarial robustness in machine learning.
\end{IEEEbiography}
\begin{IEEEbiography}[{\includegraphics[width=1in,height=1.25in,clip,keepaspectratio]{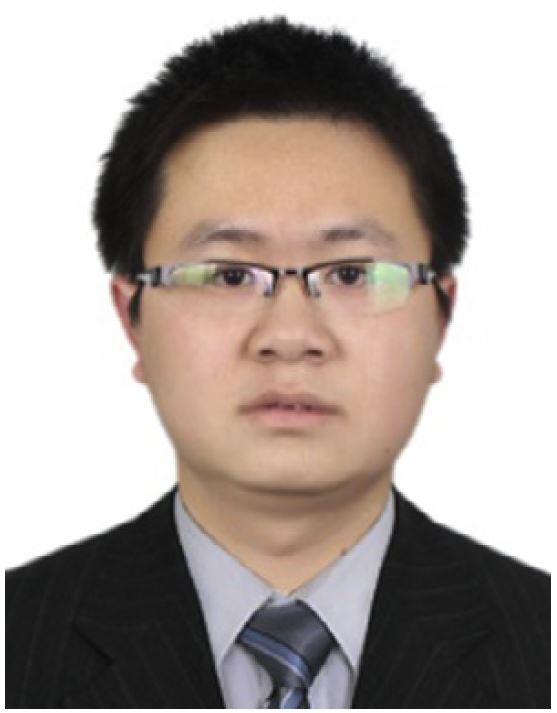}}]{Huanqian Yan} is pursuing his Ph.D. degree at
the School of Computer Science and Engineering,
Beihang University, Beijing, China. Previously, He
received his master degree in computer application
and technology from Lanzhou University in
July 2018 and his Bachelor degree in the field of
computer science and technology from Changchun
University of Science and Technology in July 2015.
His current research interests are object detection,
adversarial examples, and clustering analysis, etc.
\end{IEEEbiography}
\end{document}